\documentclass[pdflatex]{sn-jnl}

\usepackage{graphicx}%
\usepackage{multirow}%
\usepackage{amsmath,amssymb,amsfonts}%
\usepackage{amsthm}%
\usepackage{mathrsfs}%
\usepackage[title]{appendix}%
\usepackage{xcolor}%
\usepackage{textcomp}%
\usepackage{manyfoot}%
\usepackage{booktabs}%
\usepackage{placeins}%
\usepackage{tabularx}
\usepackage{algorithm}%
\usepackage{algorithmicx}%
\usepackage{algpseudocode}%
\usepackage{listings}%

\usepackage{longtable}
\usepackage{array}

\usepackage[style=apa,backend=biber]{biblatex}
\theoremstyle{thmstyleone}%
\theoremstyle{thmstyletwo}%

\theoremstyle{thmstylethree}%

\begin{document}

\title{Assessing Quality of Experience in Natural Language Generation of German Text}


\author*[1,2]{\fnm{Dinh Nam} \sur{Pham}}\email{dinh-nam.pham@campus.tu-berlin.de}

\author[2]{\fnm{Shushen} \sur{Manakhimova}}\email{shushen.manakhimova@dfki.de}

\author[2]{\fnm{Vivien} \sur{Macketanz}}\email{vivien.macketanz@dfki.de}

\author*[1,2]{\fnm{Sebastian} \sur{Möller}}\email{sebastian.moeller@tu-berlin.de}

\affil[1]{\orgname{Technische Universität Berlin}, \city{Berlin}, \country{Germany}}

\affil[2]{\orgdiv{Speech and Language Technology Lab}, \orgname{German Research Center for Artificial Intelligence (DFKI)}, \city{Berlin}, \country{Germany}}

\abstract{The rapid advancement of Natural Language Generation (NLG) has made the reliable evaluation of generated text increasingly critical, as these systems, such as large language models (LLMs), are now widely deployed in real-world applications. However, traditional automatic metrics fail to capture the multifaceted nature of perceived quality. In this paper, we introduce \texttt{TextQ-German}, a novel dataset suite for human-centered evaluation of German NLG from a Quality of Experience (QoE) perspective, covering automatic text summarization and machine translation. Through crowdsourcing studies with German speakers, we collect human quality ratings and identify relevant perceptual quality dimensions for each task. We develop automatic QoE prediction models, including transformer-based, linguistic feature-based, and hybrid approaches. Hybrid models outperform pure transformer baselines in almost all experimental settings, while linguistic features alone can approach the performance of fine-tuned language models. The dataset is extended with LLM-generated outputs annotated with overall QoE scores. Final validation on held-out sets indicates generalization to unseen data. Our work contributes a publicly accessible resource for NLG evaluation and baselines for automatic QoE prediction, providing a foundation for developing NLG systems that better align with human quality perception.
}

\keywords{Quality of Experience, Natural Language Generation, Text Quality, Readability, Text Generation, Machine Translation}

\maketitle

\section{Introduction}\label{sec1:Introduction}

In recent years, automatically generated text has become ubiquitous across a wide range of domains, from news summarization and customer service chatbots to text translation and personalized content creation. This rapid development has been driven by significant advances in Natural Language Generation (NLG), the subfield of artificial intelligence concerned with the automatic production of human-readable text. The scope of NLG is broad, with common tasks including machine translation, automatic text summarization, question-answering, and content generation, among others. These technologies promise to enhance communication, increase accessibility, and automate processes in unprecedented ways.

However, the quality of outputs from NLG systems can differ greatly, making their reliable evaluation a critical challenge. Evaluation of NLG most commonly relies on automatic metrics such as BLEU~\parencite{papineni-etal-2002-bleu} or ROUGE~\parencite{lin-2004-rouge}, which measure lexical overlap with reference texts. While these methods require no training of models and are thus convenient and computationally efficient, these metrics are known to suffer from significant limitations. As they neglect sentence structure and semantic context, they are prone to significant shortcomings in tasks that require nuanced understanding and reasoning. Most importantly, these automatic metrics correlate poorly with human judgment~\parencite{belz-reiter-2006-comparing,novikova-etal-2017-need}. The reliance on surface-level comparisons may miss the fundamental objective of whether the generated text is useful and satisfactory for its intended audience and the users of NLG systems.

As NLG systems increasingly serve as interfaces for real-world users, we argue that their ultimate success hinges not on a technical score but on the Quality of Experience (QoE) they provide. QoE, a concept well-established in the domain of telecommunications and multimedia, is defined as the degree of user delight or annoyance with an application or service. It encompasses the user's entire subjective perception, including expectations, context, and emotional response. Hence, the quality of a machine-generated text can be evaluated from the perspective of its end-user in addition to objective metrics.

We therefore aim to assess the QoE of NLG specifically for the German language, a domain that has received comparatively less attention than the English language in NLG evaluation research. To this end, we provide two key contributions: (1) a novel QoE dataset of human-annotated ratings for machine-generated outputs and (2) a suite of automatic prediction models trained on these ratings to serve as robust QoE evaluators for unseen text. We focus on two established NLG tasks, Automatic Text Summarization (ATS) and Machine Translation (MT), and investigate QoE at two complementary levels: fine-grained perceptual quality dimensions and overall perceived quality. A core part of our work involves identifying the perceptual dimensions most relevant to users and developing meaningful ways to evaluate them. Rather than pre-specifying a fixed set of quality criteria, we derive task-specific dimensions from user ratings and investigate how these dimensions can be modeled automatically. Moreover, rather than relying exclusively on learned text representations, we additionally investigate which measurable linguistic properties are associated with perceived quality and whether these features can complement transformer-based representations. By grounding our evaluation in a user-centric framework, we move beyond traditional metrics and offer a more human-oriented benchmark for the assessment of modern NLG systems. 

This article consolidates and substantially extends two previous publications from the same research project. Section~\ref{sec3.1:original-corpora} covers~\textcite{Manakhimova2025_1255}, which introduced the original ATS and MT corpora and identified and validated their perceptual quality dimensions, while the initial dimension-level prediction experiments of Section~\ref{sec4:transformer_models} build on~\textcite{pham-etal-2025-modeling}. The remaining dataset subsets, model approaches, validation experiments, and linguistic features are introduced in the present work. The datasets and additional data are publicly accessible at \url{https://github.com/DFKI-NLP/TextQ/}.

\section{Related Work}\label{sec2:Related}

\paragraph{Human-centered evaluation of NLG}
The evaluation of Natural Language Generation (NLG) has traditionally relied on a combination of human judgments and automatic metrics. Reference-based metrics such as BLEU \parencite{papineni-etal-2002-bleu} and ROUGE \parencite{lin-2004-rouge} provide efficient and reproducible measures. However, their ability to represent human-perceived quality is limited. In particular, automatic word-overlap metrics have been shown to correlate weakly or inconsistently with human ratings of generated text \parencite{belz-reiter-2006-comparing,novikova-etal-2017-need}. On the other hand, learned metrics incorporate contextual representations or are trained directly on human judgments, as exemplified by BERTScore \parencite{Zhang2020-bertscore} and COMET \parencite{rei-etal-2020-comet}.

Human evaluation remains important because text quality is inherently multifaceted. Depending on the NLG task, assessments consider properties such as fluency, adequacy, coherence, consistency, relevance, and overall quality. At the same time, the terminology and experimental procedures used for these dimensions vary considerably across studies. \textcite{van-der-lee-etal-2019-best} provide recommendations for conducting human evaluations of generated text, while \textcite{howcroft-etal-2020-twenty} identify substantial inconsistencies in the definitions of quality criteria used throughout the NLG literature. Similar concerns persist in different tasks. For automatic text summarization, SummEval \parencite{fabbri-etal-2021-summeval} evaluates summaries along coherence, consistency, fluency, and relevance and demonstrates that automatic metrics capture these dimensions to different degrees. For machine translation, the choice of annotators and evaluation protocol can substantially affect human system rankings \parencite{freitag-etal-2021-experts}. More recently, LLM-based evaluators such as G-Eval \parencite{liu-etal-2023-g} have been proposed to approximate multidimensional human evaluation. These developments motivate evaluation approaches that are explicitly grounded in human perception rather than solely in task-specific automatic scores.

\paragraph{Quality of Experience and subjective quality prediction}
QoE offers a user-centered perspective as it describes the subjective quality experienced by a user and is influenced not only by technical properties of a system but also by the content, context, and characteristics and expectations of the user \parencite{itu-p10,Moller2014-so}. Consequently, subjective experiments require the quantification of QoE assessment, often resulting in Mean Opinion Scores (MOS) or ratings of multiple perceptual quality dimensions. These subjective judgments can serve as targets for models that estimate perceived quality automatically. This paradigm is well established for speech, audio, image, and video services. For instance, subjective video-quality methodologies are standardized in ITU-T P.910, while objective models are designed to predict human-perceived quality from measurable characteristics \parencite{itu-p910}.

Only recently, however, has this same perspective been turned toward text. \textcite{naderi2019-readability} formulated German text readability as a prediction problem, demonstrating that subjective perceptions of textual complexity can be modeled automatically. Moving beyond a single property such as readability to capture the overall perception of machine-generated text, our dataset includes human ratings for both overall QoE and individual attributes, enabling us to empirically derive the perceptual dimensions most relevant to ATS and MT.

\paragraph{German text quality and automatic quality estimation}
Related research on German text has primarily addressed readability and complexity. Subjective German text complexity datasets have enabled the development of models that predict readers' perceived difficulty from linguistic characteristics and contextual representations \parencite{Naderi2019SubjectiveAO,seiffe-etal-2022-subjective}. The GermEval 2022 shared task further established German text complexity prediction as a regression problem, with participating approaches combining transformer representations and manually derived textual statistics (e.g., average sentence length) \parencite{salar-mohtaj-babak-naderi-2022-overview,anschutz-groh-2022-tum}. These studies demonstrate that measurable linguistic properties, such as lexical and syntactic features, can provide useful information about how readers perceive a text. However, these works focus on a single aspect of perception, namely text complexity or readability, rather than on the broader set of factors that shape the overall quality experienced by users. In contrast, our work considers a broader spectrum of perceived quality across multiple QoE dimensions and addresses a gap in the automatic evaluation of perceived quality for German NLG. Rather than predicting a single predefined quality criterion, we model multiple empirically derived QoE dimensions together with overall perceived quality across both ATS and MT. We further combine transformer-based representations with a broader set of interpretable linguistic features and evaluate the resulting models on independently collected data spanning both conventional and LLM-based NLG systems. In this way, our work extends existing German text evaluation from individual aspects such as readability or complexity toward a broader, user-centered prediction of perceived NLG quality.

\section{Dataset Resource}\label{sec3:Dataset}

QoE assessment of NLG requires data that reliably captures how users actually perceive generated text. By quantifying the satisfaction of past users, we can develop models that predict the QoE of future samples. To this end, we created and released \texttt{TextQ-German}, a dataset suite for Quality of Experience assessment of German natural language generation.

The resource is designed for three main purposes. First, it supports the analysis of perceptual quality dimensions in generated German text, which we identified through user experiments. Second, it enables the development and comparison of automatic QoE prediction models, including linguistic-feature-based models, transformer-based models, and hybrid approaches. Third, it provides LLM-based extensions and final validation sets to test whether QoE predictors remain robust beyond non-LLM-based neural text generation.

The resource covers two established NLG tasks: automatic text summarization (ATS) and machine translation (MT). It comprises six subsets: the initial ATS and MT corpora with dimension-level QoE ratings, LLM-generated extensions created after initial prediction experiments, and final validation sets for external model evaluation. Together, these subsets allow us to study both fine-grained perceptual dimensions and overall quality scores across classical and LLM-based generation scenarios. Specifically, \texttt{TextQ-ATS} and \texttt{TextQ-MT} contain the original ATS and MT corpora with dimension-level QoE ratings. \texttt{TextQ-ATS-LLM} and \texttt{TextQ-MT-LLM} extend these corpora with outputs generated by large language models and annotated with an overall QoE score. Finally, \texttt{TextQ-ATS-Val} and \texttt{TextQ-MT-Val} serve as validation sets for assessing model generalization on previously unseen data.

Section~\ref{sec3.1:original-corpora} describes the collection and creation of \texttt{TextQ-ATS} and \texttt{TextQ-MT}, including data sources, the identified QoE quality dimensions, and the curation process. Section~\ref{sec3.2:llm-corpora} then explains the LLM-based datasets \texttt{TextQ-ATS-LLM} and \texttt{TextQ-MT-LLM}, which were additionally used after the initial experiments based on \texttt{TextQ-ATS} and \texttt{TextQ-MT}. Section~\ref{sec3.3:validation} introduces the two held-out final validation sets, \texttt{TextQ-ATS-Val} and \texttt{TextQ-MT-Val}. Finally, Table~\ref{tab:textq_german_overview} provides an overview of all subsets of \texttt{TextQ-German}.

\subsection{ATS and MT Corpora} \label{sec3.1:original-corpora}
To determine which specific perceptual features, here called quality dimensions, drive users' QoE when judging MT and ATS outputs, we performed experimental crowdsourcing studies with German-speaking participants, allowing us to both identify, validate, and quantify these dimensions.

\subsubsection{Text Sources}
The ATS and MT corpora (\texttt{TextQ-ATS} and \texttt{TextQ-MT}) draw text from distinct sources with a broad range of vocabulary and topics. For automatic text summarization, German source texts were obtained from the GeWiki corpus~\parencite{frefel-2020-gewiki}, which was also used in GermEval 2020 Task 3: 2nd German Text Summarization Challenge~\parencite{Frefel2020}. These summaries were kindly made available to us by the organizers for use in this research. To ensure a broad range of quality levels, we additionally generated summaries internally. Summaries from both sources were produced using a range of extractive and abstractive approaches. Extractive methods included Lead-3~\parencite{Lead3} and TextRank~\parencite{textrank} while abstractive methods included Pointer-Generator~\parencite{pointer-generator}, Transformer~\parencite{Vaswani2017},  Convolutional Self-Attention Networks~\parencite{yang-etal-2019-convolutional}, and BERT-Transformer~\parencite{devlin-etal-2019-bert}. This resulted in diverse outputs that vary in fluency, content coverage, and coherence. The selected ATS outputs span summaries of up to five sentences.

For machine translation, we sampled English-German translations from the News Translation Task of the Conference on Machine Translation 2019 (WMT19)~\parencite{barrault-etal-2019-findings}, deliberately sampling outputs from top-ranked, mid-ranked, and bottom-ranked systems to capture a wide quality variety. 

To ensure a diverse range of quality levels and error patterns across both NLG tasks, we performed a simple error-type annotation on the summaries and translation outputs, conducted by several linguists. Based on this, we selectively included data points exhibiting various error categories (e.g., morphological, syntactic, lexical) and severities. For ATS, we also made sure to include summaries of different lengths with up to five sentences. This resulted in well-balanced corpora for both tasks, containing samples with varying levels of correctness and length. Both subsets were subsequently used in crowdsourcing experiments to collect human QoE ratings, as described next.

\subsubsection{Quality Dimension Identification}
To identify the perceptual quality dimensions that shape users' QoE for MT and ATS outputs, we ran two separate crowdsourcing studies, one for each NLG task. Both studies relied on Semantic Differential (SD) scaling, a well-established technique that uses bipolar adjective pairs as scale endpoints to capture subjective perceptions~\parencite{Osgood1958}.

\paragraph{Polar Adjective Pairs}
Before launching the main experiments, we assembled two custom sets of polar adjective pairs since the linguistic properties of translations and summaries differ substantially. Each pair comprised a positive adjective and its direct negative counterpart (e.g., \textit{simple--complicated}). Drawing on our earlier error type analyses of the MT and ATS corpora, we compiled a list of adjective pairs that reflected the most common error patterns in each text type. Linguists assisted in the selection to ensure the pairs captured relevant linguistic distinctions. The initial pool contained roughly 40 adjective pairs per text task, with partial overlap between the two sets. We aimed to cover as many linguistic facets of the texts as possible.

We then ran a small-scale pre-study, asking participants to evaluate texts from our corpora using all candidate adjective pairs. In a second step, participants rated how helpful they found each pair for the evaluation task. Based on these results, we reduced each set to approximately 20 adjective pairs, which were subsequently used in the main crowdsourcing studies.

\paragraph{Experiment Setup}
In the main experiments, participants were shown translations or summaries and instructed to assess the language quality using the adjective sets. For each bipolar pair, the two adjectives represented the endpoints on a 7-point Likert scale from 0 to 6. Participants were told to disregard text content as much as possible, acknowledging that content and language cannot always be fully separated, and were not informed that the texts were machine-generated. They were only told that texts might contain errors.

The survey flow consisted of four stages: (1) An introductory section explaining the setup and providing an example of a bipolar adjective pair. (2) A practice text, which served two purposes: familiarizing participants with the task and checking attention. The practice texts were clearly high or low in quality. If a participant's rating did not match the expected direction, they were asked to reconsider. This created a mild "being observed" effect, which was shown to improve performance and reliability of participants' responses in crowdsourcing studies~\parencite{Naderi2015}. (3) The main evaluation stage, where each text appeared separately and had to be rated on all adjective pairs before advancing. A slider was used to select an integer between 0 and 6 per pair. Each participant rated three texts. (4) A final optional section to provide feedback on the survey.

The study ran on the Crowdee platform~\parencite{Naderi2014Crowdee}, a mobile-friendly crowdsourcing system with full German localization. Eligible participants were self-identified native German speakers residing in the DACH region (Germany, Austria, Switzerland). They could take part up to five times, receiving different survey versions on repeat participation. Expected completion time was approximately 10 minutes. We generated around 15 survey variants per text task, covering 45 translations and 40 summaries in total. These item counts refer specifically to the quality-dimension identification experiment and do not represent the final corpus sizes. The subsequent quantification study in Section \ref{sec3.1.3-dimension-quantification} evaluated additional text items. A total of 350 participants completed the MT survey, and 425 completed the ATS survey for the quality-dimension identification. 

\paragraph{Results and Analysis}
To mitigate low-quality responses common in crowdsourcing~\parencite{Naderi2015}, we performed a data cleaning procedure. We discarded ratings from participants who completed the survey in 240 seconds or less (40\% of the expected duration) or who gave identical slider values for every adjective pair of a given sentence, as these behaviors suggested insufficient effort. We further computed the Inconsistency Score (IS)~\parencite{Naderi2018motivation} based on two repeated adjective pairs per sentence, allowing us to filter out outlier responses with high variance. After cleaning, each test item retained approximately 10 to 20 valid ratings.

We then performed Exploratory Factor Analysis (EFA) separately for each text type using SPSS~\parencite{ibm2021spss}, employing Maximum Likelihood extraction and PROMAX rotation with Kaiser Normalization to allow non-orthogonal (correlated) dimensions. Several adjective pairs exhibited low communalities or cross-loadings with a difference below 0.2, indicating they were either semantically redundant or insufficiently specific. Thus, we removed them to balance statistical fit with interpretability~\parencite{waltermann2010quality}. After this reduction, a four-factor structure with eight adjective pairs emerged for each text type. Goodness-of-fit was satisfactory: for MT, Pearson's chi-squared test yielded \(p = 0.36\) (\(\chi^2 = 2.06\), \(\mathrm{df} = 2\)), while for ATS, it yielded \(p = 0.63\) (\(\chi^2 = 0.92\), \(\mathrm{df} = 2\)).

\begin{table}[h]
\centering

\caption{Loadings of adjective pairs (English translations) on factors and percentage of explained variance for Machine Translation. Adapted from \textcite{Manakhimova2025_1255}}
\begin{tabular}{lrrrr}
\toprule
\textbf{Adjective pair} & \textbf{F1} & \textbf{F2} & \textbf{F3} & \textbf{F4} \\
\midrule
unambiguous--ambiguous      & .757 &      &      &      \\
precise--vague              & .947 &      &      &      \\
complete--incomplete        & .822 &      &      &      \\
clear--chaotic              & .580 &      &      &      \\
direct--ponderous           &      & .806 &      &      \\
simple--complicated         &      & .923 &      &      \\
grammatical--ungrammatical  &      &      & .958 &      \\
neat--confusing             &      &      &      & .915 \\
\midrule
\% of variance              & 53.2 & 8.4  & 10.5 & 8.0  \\
\bottomrule
\end{tabular}
\label{tab:mt_factor_loadings}
\end{table}

\begin{table}[h]
\centering

\caption{Loadings of adjective pairs (English translations) on factors and percentage of explained variance for Automatic Text Summarization. Adapted from \textcite{Manakhimova2025_1255}}
\begin{tabular}{lrrrr}
\toprule
\textbf{Adjective pair} & \textbf{F1} & \textbf{F2} & \textbf{F3} & \textbf{F4} \\
\midrule
precise--vague              & .884 &       &      &      \\
complete--incomplete        & .942 &       &      &      \\
coherent--incoherent        & .844 &       &      &      \\
logical--illogical          & .796 &       &      &      \\
simple--complicated         &      & 1.002 &      &      \\
straightforward--complex    &      & .783  &      &      \\
unambiguous--ambiguous      &      &       & .729 &      \\
predictable--unpredictable  &      &       &      & .704 \\
\midrule
\% of variance              & 54.4 & 14.3  & 10.6 & 2.6  \\
\bottomrule
\end{tabular}
\label{tab:ats_factor_loadings}
\end{table}

Table~\ref{tab:mt_factor_loadings} and Table~\ref{tab:ats_factor_loadings} illustrate the distribution of the adjective pairs on the four factors and the explained percentage of variance for MT and ATS. Across both text types, factors F1 to F4 reflect distinct quality dimensions based on the adjective pair(s) loading onto each. For MT, F1's four loading attributes indicate \textbf{Precision}, F2's two loading attributes indicate \textbf{Complexity}, and the single attributes on F3 and F4 indicate \textbf{Grammaticality} and \textbf{Transparency}, respectively. For ATS, F1's four loading attributes indicate \textbf{Linguistic Logic}, F2's two loading attributes indicate \textbf{Complexity}, and the single attributes on F3 and F4 indicate \textbf{Clarity} and \textbf{Predictability}, respectively. Complexity is the only dimension that emerges in both tasks, with \textit{simple--complicated} as a shared indicator. Notably, the dominant F1 factors of the two tasks share two adjective pairs (\textit{precise--vague}, \textit{complete--incomplete}) but diverge in their remaining indicators: for MT, F1 additionally captures ambiguity and clarity of phrasing, whereas for ATS it is characterized by coherence and logical consistency (\textit{coherent--incoherent}, \textit{logical--illogical}). We therefore retain distinct labels, as the factors reflect task-specific interpretations of accuracy: sentence-level fidelity for translation versus discourse-level cohesion for summarization. Full lists of all polar adjective pairs, including those removed during factor analysis and the reduced sets retained for subsequent quantification, are provided in Tables~\ref{tab:mt_all_adjective_pairs} and~\ref{tab:ats_all_adjective_pairs} in the Appendix. We give an overview of the identified quality dimensions for MT and ATS, as well as typical traits we associate with them in Table~\ref{tab:quality_dimensions_characteristics}.

\begin{table}[h]
\centering
\caption{Quality dimensions and their characteristics. Adapted from \textcite{Manakhimova2025_1255}}
\small
\begin{tabular}{lll}
\toprule
\textbf{Factor} & \textbf{Quality dimension} & \textbf{Characteristics} \\
\midrule
MT--F1 
& Precision 
& \begin{tabular}[t]{@{}l@{}}clear and complete phrasing\\unambiguous meaning\end{tabular} \\
\midrule
MT \& ATS--F2 
& Complexity 
& \begin{tabular}[t]{@{}l@{}}easily comprehensible\\not circumlocutory\end{tabular} \\
\midrule
MT--F3 
& Grammaticality 
& \begin{tabular}[t]{@{}l@{}}correct spelling and punctuation\\no missing words\end{tabular} \\
\midrule
MT--F4 
& Transparency 
& \begin{tabular}[t]{@{}l@{}}clear and coherent\\reasonably structured\end{tabular} \\
\midrule
ATS--F1 
& Linguistic Logic 
& \begin{tabular}[t]{@{}l@{}}accurate phrasing\\cohesive\end{tabular} \\
\midrule
ATS--F3 
& Clarity 
& \begin{tabular}[t]{@{}l@{}}direct and clear language\\content easily understandable\end{tabular} \\
\midrule
ATS--F4 
& Predictability 
& \begin{tabular}[t]{@{}l@{}}logical and expected structure\\methodical and coherent\end{tabular} \\
\bottomrule
\end{tabular}
\label{tab:quality_dimensions_characteristics}
\end{table}

\subsubsection{Quality Dimension Quantification}
\label{sec3.1.3-dimension-quantification}
Following the identification of the quality dimensions, we conducted a second crowdsourcing study to validate the reduced measurement scheme and test whether a reduced set of adjective pairs could reliably capture the same underlying constructs. For each factor revealed by the EFA, we selected the adjective pair with the highest factor loading as a representative indicator of that dimension. This yielded four adjective pairs per text type.

We then correlated the outcomes of the two experiments per text type to quantify the quality dimensions and assess the agreement between the full and reduced measurement schemes. In this validation experiment, 425 participants rated MT outputs, and 120 participants rated ATS outputs. The discrepancy in participant counts again stemmed from unusable ratings, which required multiple iterations of the survey to accumulate sufficient valid responses.

The correlation analysis followed the same procedure for both text types. We first applied Grubbs's test \parencite{Grubbs1950} to detect outliers, excluding any text sample point that emerged as a significant outlier on two or more of the four factors. After this filtering step, Spearman correlation coefficients consistently fell around 0.8 for both text types, providing initial evidence of strong alignment between the two experimental rounds.

To determine whether the two sets of ratings differed significantly, we first conducted the Jarque–Bera test \parencite{JARQUE1980255}, which did not indicate significant departures from normality across the factors for either text type. We then performed Levene's test \parencite{Levene1960} to assess the equality of variances. For all factors except MT's F2 (Complexity) and ATS's F4 (Predictability), Levene's test indicated homogeneous variances. Consequently, we applied a one-factor ANOVA \parencite{fisher1925statistical} to those factors. For the remaining two factors that violated the homogeneity assumption, we used Welch's t-test \parencite{welch1947} instead.

For ATS, the analysis showed no statistically significant difference between the identification and quantification experiments for any factor. For MT, no significant differences emerged for any factor except F2 (Complexity). However, the observed difference for MT's Complexity was only 0.5 points on a 7-point scale, and the correlation between the two experimental rounds on this factor remained high. Given that we cannot definitively attribute this small discrepancy to either the reduced number of adjective pairs or the natural variation between different crowdsourcing participant pools, we consider the difference as acceptable.

These findings demonstrate that the condensed set of four adjective pairs effectively captures the essential aspects of each quality dimension while remaining consistent across the experimental rounds. A key benefit of validating these quality dimensions is that future studies can adopt simpler, more efficient evaluation designs that impose less cognitive load on participants while still yielding detailed insights. The strong correlations between the reduced adjective-pair sets and their corresponding original factor scores indicate that the condensed evaluation preserves the underlying quality dimensions for both MT and ATS. Nevertheless, the slight difference observed for MT's Complexity factor (F2) suggests that complexity is not interpreted the same for different NLG tasks. In translations, simpler language supports fluency, whereas in summaries, structural and conceptual depth help convey concise information. This points to a broader need for making quality evaluation in NLP more task-specific.

\begin{table}[h]
\centering
\caption{Excerpt from \texttt{TextQ-ATS} illustrating its structure, with mean ratings per test item and quality dimension}
\footnotesize
\begin{tabular}{p{6.2cm}ccccc}
\hline
Text Sample & Logic & Complexity & Clarity & Predictability \\
\hline
Die Bundesstrasse 75 führt von Lübeck-Travemünde bis in die Hansestadt Lübeck und ist eine der beiden Bundesstrassen Hamburg und Lübeck. & 4.0625           & 3.875    & 3.9375  & 3.5000 \\
\hline
Der Name Meschete war ursprünglich nur ein geographischer Name, der sowohl den turkvölkischen Einwanderern, der Region und der heutigen Provinz den Namen gab. Bis stammt von der alten georgischen Region Mzcheta. Der Siedlungsschwerpunkt der Mescheten war einst in Usbekistan, den Vereinigten Staaten und der Ukraine. Und so listet beispielsweise das "Metzler Lexikon Sprache" die Sprache der Mescheten unter dem Sammelbegriff "muslimische Georgier" zusammen. Weltweit leben heute etwa 40.000 in der Türkei. & 2.2703           & 1.7297     & 2.5000  & 2.0541 \\
\hline
Viktor Gustav von Strauss und Torney war ein \"osterreichischer Dichter, Kirchenlieddichter und Zeichner. Berühmtheit erlangte er als erster Botschafter der Chinesischen Volksrepublik China. & 3.5294          & 3.544     & 3.8823  & 2.7353 \\

\hline
\end{tabular}
\label{tab:dataset_textq_ATS}
\end{table}

\begin{table}[h]
\footnotesize
\centering
\caption{Excerpt from \texttt{TextQ-MT} illustrating its structure, with mean ratings per test item and quality dimension}
\begin{tabular}{p{4.8cm}cccc}
\hline

Text Sample & Precision & Complexity & Transparency & Grammaticality \\
\hline
Die Senatorin von Massachusetts, Elizabeth Warren, sagte am Samstag, sie werde nach den Zwischenwahlen einen "harten Blick" darauf werfen, für das Präsidentenamt zu kandidieren. & 2.1765         & 3.2941   & 2.8235  & 2.4706 \\
\hline
Die Labour Party wäre nicht die erste, die eine solche Idee befürwortet, mit der grünen Party, die eine vier-Tage-Arbeitswoche während ihrer 2017 allgemeinen Wahlkampagne geschworen hat. & 3.3947 & 4.1579 & 4.1316 & 3.5526 \\
\hline
Angesichts der Erde, die schütteln, dass Indonesien ständig übersteht, bleibt das Land kläglich unvorbereitet für den Zorn der Natur. & 3.5333 & 3.8667 & 4.0000& 4.2000 \\
\hline
\end{tabular}
\label{tab:dataset_textq_MT}
\end{table}

Based on the crowdsourcing studies, we curated the evaluated text items and their human QoE dimension ratings into two datasets, \texttt{TextQ-ATS} and \texttt{TextQ-MT}. The former comprises 91 text samples, the latter contains 106. Prioritizing reliability over quantity, each sample received ratings from 10 to 20 annotators, and we averaged these per item. Tables \ref{tab:dataset_textq_ATS} and \ref{tab:dataset_textq_MT} provide example excerpts from \texttt{TextQ-ATS} and \texttt{TextQ-MT}, respectively. These subsets can be used to develop QoE prediction models.

\subsection{LLM-Based Corpora Extensions} \label{sec3.2:llm-corpora}
After initial experiments on \texttt{TextQ-ATS} and \texttt{TextQ-MT} as described in Section \ref{sec4:dimension_prediction}, we conducted crowdsourcing runs again to extend the datasets. Following the recent rapid trends in NLG and NLP technology, the text samples were now generated using LLMs. The source texts were drawn from the same datasets as the original corpora (Section~\ref{sec3.1:original-corpora}), namely the GeWiki corpus for ATS and the WMT19 English-German test set for MT; however, the samples used for this extension were different from the original corpus. In addition to the four quality dimensions, an overall QoE score was collected. The extension of the corpus was generated using a mix of commercial API-based models and locally hosted open-weight models. The OpenAI models (GPT-4o and GPT-3.5 Turbo) were accessed through the Chat Completions API, while the open-weight models were hosted locally through Ollama. The full prompts used for each model are listed in Appendix~\ref{app:prompts}.

Translations were produced with four models: the two OpenAI models, GPT-4o and
GPT-3.5 Turbo, and two open-weight models, Llama~3.2 (3B) and StableLM~2 (1.6B). All four produced translations for the complete set of source items, using a single prompt to translate the input into German. The maximum output length was capped at 300 tokens to keep the generated translations comparable in length to those in the original corpus and to avoid overly long texts that would burden crowdworkers during rating. A temperature of 0.2 was applied uniformly across all models; lower or higher temperatures caused some models to produce mixed German-English output, and a single setting was kept constant across models for consistency.

Summaries were generated with a broader set of models. From OpenAI, GPT-4o was used in a length-constrained configuration (referred to as \texttt{gpt-4o\_short}), with a temperature of 0.0, a maximum of 200 output tokens, and a prompt that explicitly restricted the summary length. The open-weight summarization models comprised DeepSeek-R1 (1.5B), StableLM~2 (1.6B), Llama~3.2 (3B), SmolLM2-German-Instruct (360M, Q8), and SauerkrautLM-7B-v1 (GGUF, Q2\_K), all hosted locally through Ollama.
These locally hosted models used a common German summarization prompt, a temperature of 0.5, and a maximum output length of 540 tokens; source texts exceeding 2,800 characters were truncated to their first 2,800 characters before insertion. As an instruction-tuned model, SauerkrautLM-7B-v1 instead used the Alpaca-style instruction format specified in its prompt template.

The generated outputs were manually annotated for quality and error types following the same general procedure used for the original corpora. Based on this annotation, outputs were selected to cover a broad range of quality levels rather than being dominated by fluent, high-quality generations. For MT, all four models produced outputs for every source item, allowing balanced selection across systems and quality levels. For ATS, the open-weight models did not all complete every source item, so the final selection was balanced across quality categories as far as the available outputs allowed. The selected texts were then evaluated in additional crowdsourcing runs using the same four dimension-specific adjective pairs described in Section~\ref{sec3.1:original-corpora}.

The overall quality score was collected as a separate rating. Alongside the dimension-specific adjective pairs, raters judged each text using an overall good--bad (\textit{gut--schlecht}) item on the same 0--6 scale, where 0 represented bad quality, and 6 represented good quality.

We refer to these additional subsets as \texttt{TextQ-ATS-LLM} and \texttt{TextQ-MT-LLM}. \texttt{TextQ-ATS-LLM} and \texttt{TextQ-MT-LLM} each contain 77 rated text samples.

\subsection{Final Validation} %
\label{sec3.3:validation}
In the same fashion as for \texttt{TextQ-ATS}, \texttt{TextQ-MT}, \texttt{TextQ-ATS-LLM}, and \texttt{TextQ-MT-LLM}, we obtained additional samples for ATS and MT as held-out final validation sets, referred to as \texttt{TextQ-ATS-Val} and \texttt{TextQ-MT-Val}. These sets were drawn from the same generation effort as the LLM-based extensions: from the annotated output pool, we made a separate manual selection for validation. Source texts did not overlap with those used for any other \texttt{TextQ-German} subset, and selection was again balanced across the quality range. As a result, each validation set combines items produced by the LLMs described in Section~\ref{sec3.2:llm-corpora} with items produced by the non-LLM neural systems introduced in Section~\ref{sec3.1:original-corpora}, yielding a heterogeneous mix of generation systems. For \texttt{TextQ-ATS-Val}, the non-LLM items were mostly generated by two of these systems, BERT-Transformer and Convolutional Self-Attention Networks; for \texttt{TextQ-MT-Val}, the non-LLM items were drawn from the WMT19 submissions. \texttt{TextQ-ATS-Val} comprises 77 items and \texttt{TextQ-MT-Val} 76, each split approximately evenly between LLM-generated and non-LLM items. All items are annotated with the four quality dimension ratings and the overall QoE score. Each item carries the same labels as the LLM-based subsets---the four quality dimensions and the overall QoE score, collected in the same way. These two subsets serve as held-out validation sets for the final evaluation of the best QoE predictors developed for ATS and MT.

\begin{table*}[t]
\caption{Overview of the \textsc{TextQ-German} dataset suite. ATS denotes automatic text summarization, MT denotes machine translation. All labels are represented as mean rating on a 0--6 scale.}
\centering
\footnotesize
\setlength{\tabcolsep}{4pt}
\renewcommand{\arraystretch}{1.15}
\begin{tabularx}{\textwidth}{@{}l l l c X X@{}}
\toprule
\textbf{Subset} 
& \textbf{Task} & \textbf{Generation} & \textbf{Size} & \textbf{QoE labels} & \textbf{Purpose} \\
\midrule

\texttt{TextQ-ATS} & ATS & Non-LLM & 91 & 4 dimensions & Dimension-level QoE prediction \\

\texttt{TextQ-MT} & MT & Non-LLM & 106 & 4 dimensions & Dimension-level QoE prediction \\

\hline

\texttt{TextQ-ATS-LLM} & ATS & LLM & 77 & Overall score and 4 dimensions & Overall QoE prediction and dimension-level QoE data extension \\

\texttt{TextQ-MT-LLM} & MT & LLM & 77 & Overall score and 4 dimensions & Overall QoE prediction and dimension-level QoE data extension \\

\hline

\texttt{TextQ-ATS-Val} & ATS & LLM and Non-LLM & 77 & Overall score and 4 dimensions & Held-out final validation \\

\texttt{TextQ-MT-Val} & MT & LLM and Non-LLM & 76 & Overall score and 4 dimensions & Held-out final validation \\
\bottomrule
\end{tabularx}
\label{tab:textq_german_overview}
\end{table*}

Table \ref{tab:textq_german_overview} gives an overview of the datasets. The datasets of \texttt{TextQ-German} can be accessed at \url{https://github.com/DFKI-NLP/TextQ/}. We make them publicly available under a CC BY-NC 4.0 license for non-commercial purposes.

\section{Methodology}\label{sec4:Experiments}
The \texttt{TextQ-German} datasets feature scaled human ratings of text samples, offering a valuable resource for training and developing models that predict ratings for new, unseen samples, thus facilitating automatic QoE assessment. This approach of training models on human scores and subsequently deploying them as predictors of subjective perceptual quality has been successfully applied beyond text to other media types, including video \parencite{Li2019VSFA}, image \parencite{ma2025surveyimagequalityassessment}, and speech \parencite{wang-etal-2025-qualispeech}. Following this, we train and evaluate models on our \texttt{TextQ-German} datasets for machine-generated text (MT and ATS). This section details the models and experiments that we conducted to this end.

We consider two levels of prediction. The first is dimension-level QoE prediction, where models estimate each of the perceptual dimensions identified in the crowdsourcing studies. The second is overall QoE prediction, where models estimate a single quality score for a generated text.

The experimental design follows the structure of \texttt{TextQ-German}. We first use the initial corpora, \texttt{TextQ-ATS} and \texttt{TextQ-MT}, to study dimension-level QoE prediction. For ATS, the target dimensions are linguistic logic, complexity, clarity, and predictability. For MT, the target dimensions are precision, complexity, transparency, and grammaticality. These experiments test whether fine-grained perceptual quality dimensions can be predicted from textual information. We then move to the overall QoE prediction. This setting reflects practical evaluation scenarios in which a single quality estimate is needed. For this purpose, we use the subsets that contain overall QoE ratings, namely \texttt{TextQ-ATS-LLM} and \texttt{TextQ-MT-LLM}. 

The experiments are organized into four stages. First, we benchmark models for predicting the four QoE dimensions of ATS and MT. Second, we train models for predicting the overall QoE score. Third, we extend the experiments to also cover LLM-generated data. Finally, we evaluate the best-performing configurations on the held-out validation sets, \texttt{TextQ-ATS-Val} and \texttt{TextQ-MT-Val}. These final validation sets are not used for feature selection, hyperparameter tuning, checkpoint selection, or model selection.

All prediction tasks are formulated as regression problems. Individual participants provide discrete ratings on a 0–6 scale, but the prediction targets are item-level Mean Opinion Scores (MOS), obtained by averaging the ratings across annotators. These aggregated scores are therefore continuous-valued. For dimension-level prediction, we consider both single-target settings, where one model is trained for each quality dimension, and multi-target settings, where one model predicts all dimensions jointly. For overall QoE prediction, models output a single continuous score.

\subsection{Models} \label{sec4:models}
In this section, we describe the different models we used for the experiments to predict QoE.

\subsubsection{Transformer-Based Models} \label{sec4:transformer_models}
The size of our datasets is relatively limited for training purposes, given that human annotation is a time- and resource-demanding process. Hence, as an initial baseline, we chose to fine-tune pre-trained German language models, taking advantage of their pre-existing knowledge of German text.

More specifically, we fine-tuned five pre-trained transformer-based models derived from Huggingface\footnote{\url{https://huggingface.co/}}:
\begin{itemize}
\item \texttt{bert-base-german-uncased} \parencite{bert-based-uncased}
\item \texttt{bert-base-german-cased} \parencite{bert-base-cased}
\item \texttt{gbert-base} \parencite{chan-etal-2020-germans}
\item \texttt{gbert-large} \parencite{chan-etal-2020-germans}
\item \texttt{gelectra-large} \parencite{chan-etal-2020-germans}
\end{itemize}
For each model, we replaced the output layer with a linear layer. Its output dimensionality was set to 1 when the goal was to predict either a single QoE dimension or an overall score, and to 4 when the goal was to perform simultaneous multi-target regression across all four quality dimensions.

\subsubsection{Linguistic Feature-Based Models} \label{sec4:feature_models}
Beyond the task of rating prediction, gaining insight into which linguistic features or statistical text properties can be used to estimate text QoE is of significant value. Thus, we employed feature selection (FS) methods to identify the linguistic features most relevant to predicting perceived text quality. We then assessed the performance of models that rely exclusively on these selected features.

To this end, we implemented 121 textual features in an effort to facilitate a comprehensive search for relevant features for the text quality dimensions. Every feature is a real-valued number computable for a given text, derived from the established literature and areas such as complexity and readability assessment. To compute these features, we used the spaCy library \parencite{ines_montani_2023_10009823} with its German pipeline for tokenisation, part‑of‑speech tagging, and dependency parsing. For readability indices and lexical diversity, we relied on standard implementations from the textstat library \parencite{textstat}, supplemented with custom functions for German‑specific measures.
The features can be categorized in the following feature types: readability features, lexical richness features, syntactic features, and morphological features. A full list of all features implemented can be found in Table \ref{tab:all-features} in the Appendix. In the following, we list some of the feature types with specific examples of the features we included in the experiments:

\begin{itemize}
    \item \textbf{Traditional Readability Features}: Flesch Reading Ease, Average sentence length in words, Average word length in characters, Wiener Sachtextformel, Gunning fog index, SMOG index, Coleman-Liau index, \ldots
    \item \textbf{Lexical Richness Features}: Dugast's Uber Index, Type-Token-Ratio, Lexical density, Noun variation, Modifier variation, Adverb ratio, \ldots
    \item \textbf{Syntactic Features}: Noun phrases per sentence count, Verb phrases per sentence count, \ldots
    \item \textbf{Inflectional Morphology of the Verb}: Infinitive-verbs-to-verbs-ratio, Participle-verbs-to-verbs-ratio, Past-tense-verbs-to-finite-verbs-ratio, third-person-to-finite-verbs-ratio, \ldots
    \item \textbf{Inflectional Morphology of the Noun}: Genitive-nouns-to-nouns-ratio, accusative-nouns-to-nouns-ratio, dative-nouns-to-nouns-ratio, \ldots
\end{itemize}

These features were computed for all text samples of the datasets for the experiments. Using these features and the QoE dimension ratings as labels, features can be selected to fit models, such as a linear regression model.

Given the number of 121 linguistic features that we implemented and can compute for a given text, the total number of possible feature subsets is $2^{121}-1$, making a brute-force search for the optimal feature set computationally infeasible. However, selecting an optimal subset serves our two objectives: (a) identifying what linguistic features are relevant for assessing the text quality and (b) improving future ML models' performance by minimizing redundancy within the set of selected features. Therefore, we approach this as the process of feature selection (FS) and utilize commonly used FS techniques that choose the most important features.

We implemented the FS methods Recursive Feature Elimination (RFE) and Sequential Feature Selection (SFS) using linear regression as the base model. This choice was motivated by the computational efficiency, high interpretability due to the linear equation as well as comparatively higher robustness to overfitting. As SFS requires a number $n$ of features to select, we set $n$ to 20 and used forward selection.
Additionally, the FS techniques Lasso and Elastic net were also employed, which combine feature selection with model training, complementing the wrapper methods with an embedded approach. 

In summary, for any given text, we compute 121 real-valued linguistic features. Using the methods described above, we can fit a predictive model and obtain label predictions while simultaneously gaining insight into which individual features were selected.

\subsubsection{Hybrid Models} \label{sec4:hybrid_models}
Beyond using linguistic features or pre-trained language models in isolation, we propose hybrid models that combine the learned representations of language models with hand-crafted linguistic features. Jointly making use of both sources of information allows the model to benefit from the rich, context-aware representations captured by transformers while also incorporating explicitly defined linguistic knowledge, such as readability, lexical diversity, and syntactic complexity.

We constructed two types of hybrid models as follows.

\begin{itemize}
    \item \textbf{Hybrid Language Model}: Using the training set, a FS method first selects a set of linguistic features. For each input text, we compute this feature set while simultaneously forwarding the text through a language model backbone to extract its [CLS] token embedding. The [CLS] token is a special token prepended to the input sequence in transformer models like BERT. Its final hidden state aggregates contextual information from the entire text into a fixed-size vector whose dimensionality is determined by the language-model backbone. This vector serves as a compact semantic representation of the input. Both the linguistic feature set and this embedding are then concatenated and passed to a linear layer, and the entire network is fine-tuned end-to-end. This architecture integrates linguistic indicators with neural representations in a late fusion manner, allowing the model to learn complementary patterns from both sources.

    \item \textbf{Hybrid SVM}: The training procedure follows two stages. Stage 1: A language model is fine-tuned on the training set in the same way as the models in Section~\ref{sec4:transformer_models}. Simultaneously, a FS method selects a feature set for the training set. Stage 2: We extract the fine-tuned model's [CLS] embeddings for each training sample, concatenate them with the corresponding selected linguistic features, and train a support vector machine (SVM) on this combined representation. During inference, the SVM predicts QoE ratings for new samples by taking as input both the computed features and the embeddings produced by the frozen, fine-tuned language model. The language model weights remain frozen throughout evaluation. For the support vector regression model, we used the radial basis function (RBF) kernel, and set $C$ to $1.0$ and epsilon to $0.1$.
\end{itemize}

\subsubsection{Multi-Task Models} \label{sec4:multitask_models}
As an alternative to single-task models, we explored a multi-task learning (MTL) approach, motivated by evidence that MTL can improve performance in multi-target regression settings \parencite{MohtajSK023}. Our MTL architecture employs a shared pre-trained language model backbone across all tasks. All parameters of the language model are shared between the tasks, while each task has a separate task-specific linear regression head. We extract the pooled [CLS] token embedding from this shared encoder and forward it to the corresponding task-specific regression head. Each head independently predicts one target variable. During training, the losses of all tasks are summed with equal weighting, such that gradients from each task update the same shared language-model parameters. The tasks can correspond either to the QoE dimensions or to the two datasets (ATS and MT). Thus, we adopt a hard parameter sharing approach. 

\begin{figure}[!p]
    \centering
    \includegraphics[width=0.9\textwidth]{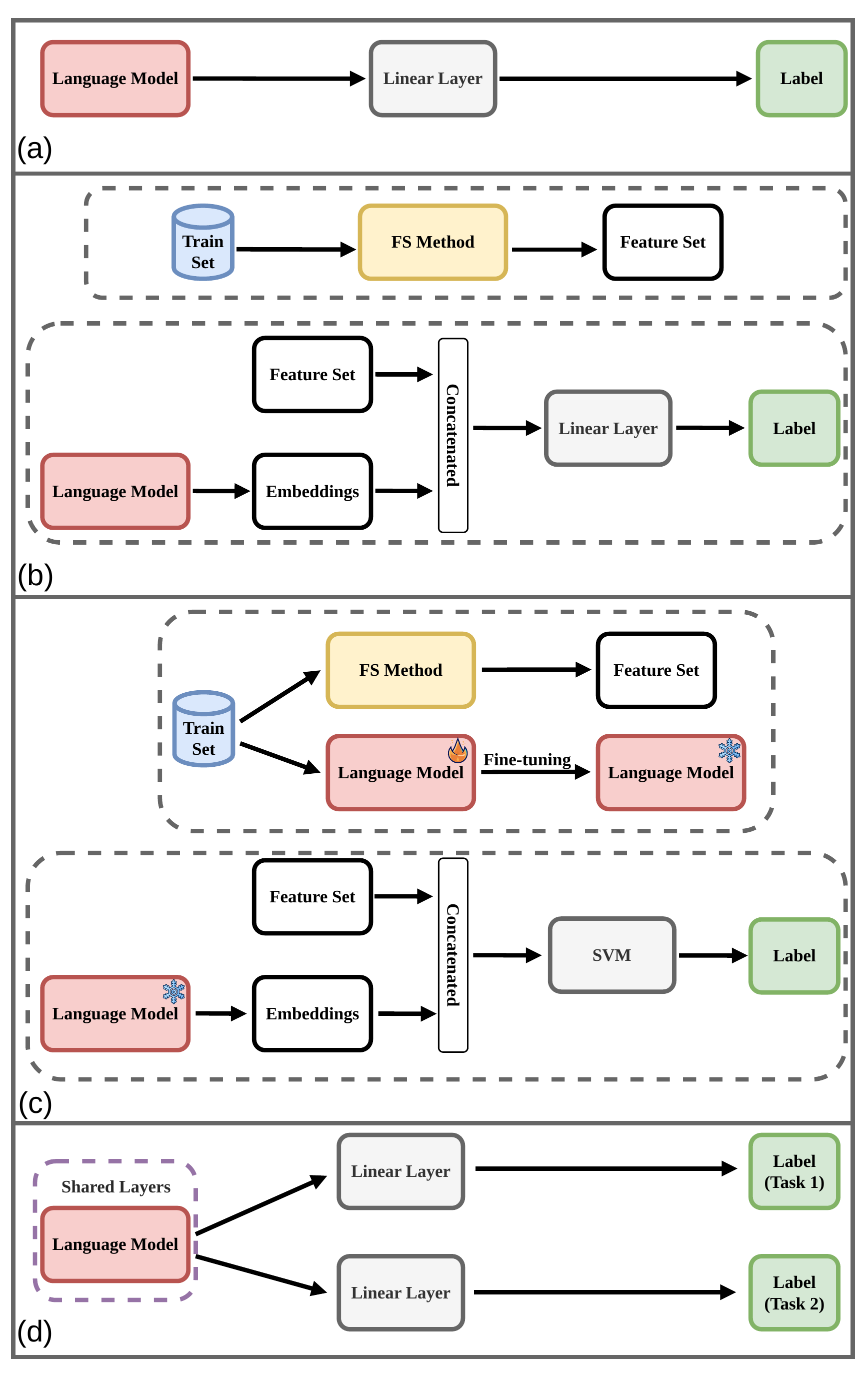}
    \caption{Illustration of (a) transformer-based models, (b) hybrid language models, (c) hybrid SVM, and (d) multi-task architecture. Dashed outlines group components belonging to the same processing or training stage.}
    \label{fig:hybrid_models}
\end{figure}

Figure \ref{fig:hybrid_models} illustrates an overview of the presented models.

\subsection{Experimental Setup}
For all experiments, we maintained consistent hyperparameters and data splits to ensure fair comparability. The chosen hyperparameters were as follows: learning rate \(2 \times 10^{-5}\), mean squared error (MSE) loss function, 30 epochs, batch size 8, and AdamW optimizer with weight decay 0.01.

Due to the relatively small size of our datasets, we employed a 7-fold cross-validation. Each dataset was randomly shuffled and then split into seven equally sized folds, denoted \(F = \{f_1, \ldots, f_7\}\). In each cross-validation round, fold \(f_i\) was designated as the test set, fold \(f_{i-1}\) (or \(f_7\) in the case of \(i = 1\)) as the validation set, and the remaining five folds as the training set. This procedure assigned each fold to both validation and test roles exactly once. By fixing random seeds, we guaranteed that all models trained on the same dataset were evaluated on the same fold partitions and sample orders.

We monitored the validation root mean squared error (RMSE) at the end of every epoch and retained the checkpoint achieving the lowest value for evaluation on the test fold. All reported metrics represent averages over the seven test folds.

In the experiments that used linguistic features, we standardized the features by subtracting the mean and scaling to unit variance based on the training data. The test fold was then normalized using the mean and standard deviation computed from the training folds to avoid data leakage. Importantly, feature selection was repeated independently in every cross-validation round and fitted exclusively on the corresponding training partition. Neither the validation fold nor the test fold was used to determine the selected linguistic features.

Unlike the neural models, the SVM does not require a validation set for early stopping. Therefore, to maximize the available training data, we fit the SVM on the union of the training and validation folds for each cross-validation round. The feature subset used by the SVM remained the subset selected from the training partition of that round.

\subsection{Dimension-Level QoE Prediction} \label{sec4:dimension_prediction}
With the models and experimental settings established, we now present our experiments for assessing QoE dimensions in NLG. We organized our evaluation around four modeling choices.

\paragraph{Language Model Selection}
We first fine-tuned and evaluated the pre-trained language models described in Section~\ref{sec4:transformer_models} on both \texttt{TextQ-MT} and \texttt{TextQ-ATS}. Each model was equipped with a regression head of four output units to predict all QoE dimensions simultaneously in a multi-target regression setup. Based on these experiments, we identified the best-performing language model for each of the two text tasks, which served as the backbone for all subsequent experiments.

\paragraph{Multi-Label Regression Approaches}
With the best language model selected, we compared three regression approaches for dimension-level QoE prediction. Each text sample has four QoE dimension values to predict, making this a multi-label regression problem. First, we retained the multi-target regression approach used before, where a single model predicts all four dimensions at once. Second, we implemented single-target regression, instantiating a separate model for each of the four dimensions with an output size of one. Third, we applied the multi-task learning (MTL) architecture described in Section~\ref{sec4:multitask_models}, treating each QoE dimension as an independent task with a shared backbone.

\paragraph{Linguistic Feature Selection}
Beyond neural approaches, we evaluated how well traditional linguistic features can predict QoE dimensions. We applied the feature selection methods described in Section~\ref{sec4:feature_models} to both datasets, assessing the predictive performance achievable with carefully selected hand-crafted features alone.

\paragraph{Hybrid Models}
Finally, we combined the best-performing language model backbone with the most effective feature selection method to train and evaluate the proposed hybrid models on both MT and ATS. These experiments allowed us to determine whether linguistic features provide a complementary predictive signal beyond what the language model learns from raw text.

\subsection{Overall QoE Prediction}
\label{sec4:overall_prediction}

Following the dimension-level experiments, we extended our evaluation to overall QoE prediction using the subsequently newly obtained LLM-generated text corpora (\texttt{TextQ-MT-LLM} and \texttt{TextQ-ATS-LLM}), each annotated with a single overall QoE score per sample.

We first re-evaluated all five pre-trained language models on both LLM datasets in a single-target regression setup, where each model outputs a single value per text sample.

Second, we investigated whether jointly training on both datasets could improve overall score prediction. Using the previously identified best-performing language model as the backbone, we implemented two strategies. The first strategy was multi-task learning, where each dataset was treated as a separate task with a shared backbone. The second strategy was simply merging both datasets into a single corpus and fine-tuning the language model on this combined data, following the standard procedure.

Finally, based on the outcomes of these experiments, we selected the optimal language model and training strategy for the hybrid models in the context of overall QoE prediction for ATS and MT.

\subsection{LLM-Based Extension}
\label{sec4:llm_training}

The newly collected LLM-generated data can also be combined with the original \texttt{TextQ-ATS} and \texttt{TextQ-MT} corpora to extend dimension-level QoE assessment. To this end, we compared three model variants using the previously selected language model backbone and feature selection method: (1) the pure language model, (2) the hybrid language model, and (3) the hybrid SVM, following the same experimental protocol as before.

\subsection{Final Evaluation}
\label{sec4:final_evaluation}

To assess generalization to completely unseen data, we conducted a final evaluation using \texttt{TextQ-MT-Val} and \texttt{TextQ-ATS-Val}. These datasets were collected via crowdsourcing after the main experiments and serve as held-out final evaluation sets, having played no role in feature selection, hyperparameter tuning, checkpoint selection, or model selection. For each text task, we applied all seven model instances obtained from the 7-fold cross-validation procedure. Each instance performed inference on all samples of the corresponding final evaluation set, and we report the average performance across the seven instances. This setup provides an estimate of how well our models generalize to new, unseen text samples.

\section{Results and Discussion}\label{sec5:Results-Discussion}
We report root mean squared error (RMSE), mean absolute error (MAE), and the coefficient of determination ($R^2$) as metrics for regression. We primarily use RMSE for model selection because it assigns greater weight to larger prediction errors, which are particularly relevant when assessing deviations from human QoE ratings. The metrics are averaged across folds and dimensions.

\subsection{Dimension-Level QoE Prediction}

\paragraph{Language Models}
The fine-tuning performance of each pre-trained language model is summarized in Tables \ref{tab:model_selection_ATS} and \ref{tab:model_selection_MT}. For the ATS dataset, \texttt{gbert-large} yielded the lowest RMSE, whereas \texttt{gelectra-large} was the best-performing model on the MT dataset. Across all models, prediction errors were systematically higher for MT than for ATS. This pattern suggests that QoE prediction may be more demanding for machine translation. Nevertheless, we acknowledge that cross-dataset differences in data quality or annotation reliability could partially explain this discrepancy.

\begin{table}[h]
\caption{Performance of the fine-tuned language models on \texttt{TextQ-ATS}}
\centering
\begin{tabular}{lcccc}
\hline
    Model & RMSE & MAE & $R^2$ \\
\hline
\texttt{bert-base-german-uncased} & 0.9751 & 0.8168 & 0.0800 \\
\texttt{bert-base-german-cased} & 0.9097 & 0.7458 & 0.2047 \\
\texttt{gbert-base} & 0.9997 & 0.8258 & 0.0549	 \\
\texttt{gbert-large} & \textbf{0.8438} & \textbf{0.6948} & \textbf{0.2927} \\
\texttt{gelectra-large} & 0.8693 & 0.7142 & 0.2657 \\
\hline
\end{tabular}
\label{tab:model_selection_ATS}
\end{table}

\begin{table}[h]
\caption{Performance of the fine-tuned language models on \texttt{TextQ-MT}}
\centering
\begin{tabular}{lcccc}
\hline
    Model & RMSE & MAE & $R^2$ \\
\hline
\texttt{bert-base-german-uncased} & 1.1449 & 0.9479 & 0.1865 \\
\texttt{bert-base-german-cased} & 1.1375 & 0.9499 & 0.1795 \\
\texttt{gbert-base} & 1.1538 & 0.9403 & 0.1716 \\
\texttt{gbert-large} & 1.0932 & 0.9221 & 0.2026 \\
\texttt{gelectra-large} & \textbf{0.9853} & \textbf{0.8350} & \textbf{0.3899} \\
\hline
\end{tabular}
\label{tab:model_selection_MT}
\end{table}

Moreover, while both MAE and RMSE share the same unit as the target variables, MAE is consistently lower than RMSE, as expected from their respective formulations \parencite{Willmott2005AdvantagesOT}. 
Based on the RMSE, we selected \texttt{gbert-large} for ATS and \texttt{gelectra-large} for MT as the backbones for all subsequent experiments.

\paragraph{Multi-Label Regression Approaches}

\begin{table}[h]
\caption{Performance of the multi-label regression approaches on \texttt{TextQ-ATS}}
\centering
\begin{tabular}{lcccc}
\hline
    Model & RMSE & MAE & $R^2$ \\
\hline
Multi-Target (\texttt{gbert-large}) & \textbf{0.8438} & \textbf{0.6948} & \textbf{0.2927} \\
Single-Target (\texttt{gbert-large}) & 0.9566 & 0.7735 & 0.1051 \\
Multi-Task Learning (\texttt{gbert-large}) & 1.0080 & 0.8389 & -0.0152	 \\
\hline
\end{tabular}
\label{tab:multilabel_ATS}
\end{table}

\begin{table}[h]
\caption{Performance of the multi-label regression approaches on \texttt{TextQ-MT}}
\centering
\begin{tabular}{lcccc}
\hline
    Model & RMSE & MAE & $R^2$ \\
\hline
Multi-Target (\texttt{gelectra-large}) & 0.9853 & 0.8350 & 0.3899 \\
Single-Target (\texttt{gelectra-large}) & \textbf{0.9657} & \textbf{0.8149} & \textbf{0.4141} \\
Multi-Task Learning (\texttt{gelectra-large}) & 1.0430 & 0.8783 & 0.3034	 \\
\hline
\end{tabular}
\label{tab:multilabel_MT}
\end{table}

Tables~\ref{tab:multilabel_ATS} and \ref{tab:multilabel_MT} show the results for multi-label regression. The multi-task model performed worst on both text tasks, though only slightly. We hypothesize that the limited amount of training data restricts the model's ability to learn complex task interactions. In our current implementation, the multi-task architecture is essentially a regressor on top of the language model embeddings. Given more data, a larger or more sophisticated prediction head might better exploit the shared representations across tasks. For ATS, the multi-target regression model achieved the lowest error, while for MT, single-target regression performed the best. We therefore retained these configurations for the remaining experiments with hybrid models.

\paragraph{Linguistic Feature Selection}

\begin{table}[!h]
\centering
\caption{Performance of feature selection methods on \texttt{TextQ-ATS}, RMSE values and feature counts are averaged across the four ATS quality dimensions}
\label{tab:features_ats_summary}
\begin{tabular}{lcc}
\toprule
Feature selection method & RMSE & Average number of selected features \\
\midrule
RFE             & 1.0523 & 1.00   \\
SFS               & \textbf{0.8923} & 20.00 \\
Lasso             & 1.0068 & 6.00   \\
ElasticNet        & 1.0482 & 7.50   \\
\bottomrule
\end{tabular}
\end{table}

\begin{table}[!h]
\centering
\caption{Performance of feature selection methods on \texttt{TextQ-MT}, RMSE values, and feature counts are averaged across the four MT quality dimensions}
\label{tab:features_mt_summary}
\begin{tabular}{lcc}
\toprule
Feature selection method & RMSE & Average number of selected features \\
\midrule
RFE          & 1.2613 & 7.00   \\
SFS               & \textbf{1.0257} & 20.00 \\
Lasso             & 1.1823 & 5.50   \\
ElasticNet        & 1.1981 & 6.75   \\
\bottomrule
\end{tabular}
\end{table}

Tables~\ref{tab:features_ats_summary} and~\ref{tab:features_mt_summary} describe the performance of the feature selection methods on QoE prediction as well as the average number of features selected out of all 121 possible linguistic features. Detailed per-dimension results and the specific selected features are available in the Appendix (Tables~\ref{tab:features_ats_detailed} and~\ref{tab:features_mt_detailed}). Across all methods, the selected feature subsets were notably small. Recursive Feature Elimination (RFE) was particularly aggressive for ATS, selecting a single feature per dimension, while selecting larger subsets for some MT dimensions. Sequential Feature Selection (SFS) consistently yielded the lowest RMSE values across both datasets and all quality dimensions, with average RMSE scores of 0.8923 (ATS) and 1.0257 (MT). These results approach those of the best-performing language models, which achieved 0.8438 and 0.9853, respectively.

The result that a simple linear regression model fitted on a relatively small number of carefully selected linguistic features can rival fine-tuned transformer-based models is striking. This finding strongly motivates our proposed hybrid approach, which aims to combine the complementary strengths of complex neural representations and hand-crafted linguistic features. Additionally, FS methods offer a practical advantage: they provide transparency by explicitly revealing which linguistic features drive QoE prediction. Given its clear superiority, we use SFS as the feature selection method for all subsequent hybrid model experiments.

\paragraph{Hybrid Models}

\begin{table}[h]
\caption{Performance of the hybrid models on \texttt{TextQ-ATS}}
\centering
\begin{tabular}{lcccc}
\hline
    Model & RMSE & MAE & $R^2$ \\
\hline
Multi-Target (\texttt{gbert-large}) & 0.8438 & 0.6948 & 0.2927 \\
Hybrid language model & 0.8557 & 0.70275 & 0.3033 \\
Hybrid SVM & \textbf{0.8254} & \textbf{0.6645} & \textbf{0.3664} \\
\hline
\end{tabular}
\label{tab:hybrid-models-ats}
\end{table}

With the best-performing language model backbone, multi-label regression approach, and feature selection method established for each text task, we evaluated the proposed hybrid models. For both tasks, SFS selected 20 linguistic features, while the respective large language-model backbones produced 1,024-dimensional [CLS] embeddings. Consequently, the concatenated representation used by the hybrid models comprised 1,044 features. Tables~\ref{tab:hybrid-models-ats} and~\ref{tab:hybrid-models-mt} report the performance of the hybrid language model and hybrid SVM against the respective neural baselines.

\begin{table}[h]
\caption{Performance of the hybrid models on \texttt{TextQ-MT}}
\centering
\begin{tabular}{lcccc}
\hline
    Model & RMSE & MAE & $R^2$ \\
\hline
Single-Target (\texttt{gelectra-large}) & 0.9657 & 0.8149 & 0.4141 \\
Hybrid language model & \textbf{0.9127} & \textbf{0.7510} & \textbf{0.4720} \\
Hybrid SVM & 0.9584 & 0.8091 & 0.4233 \\
\hline
\end{tabular}
\label{tab:hybrid-models-mt}
\end{table}

On the ATS dataset, the hybrid SVM achieved the best overall performance, reducing RMSE from 0.8438 to 0.8254 and increasing \(R^2\) from 0.2927 to 0.3664. On the MT dataset, the hybrid language model performed the best, lowering RMSE from 0.9657 to 0.9127 and raising \(R^2\) from 0.4141 to 0.4720.

These results clearly demonstrate the merit of combining neural embeddings with carefully selected linguistic features. On both datasets, at least one hybrid model outperforms the pure transformer-based baseline, showing that linguistic features provide complementary predictive information not fully captured by the language model alone. Interestingly, the optimal hybrid architecture differs by text type. For ATS, which involves multi-sentence summaries, the SVM-based hybrid excels, perhaps because the feature set captures text-level properties such as lexical diversity and readability that are particularly salient for longer texts. For MT, where inputs are mostly single sentences, the neural hybrid performs better, suggesting that contextualized embeddings are especially powerful for sentence-level tasks. Overall, these findings validate that linguistic features and neural representations encode complementary information, and their combination yields measurable improvements in QoE prediction.

\subsection{Overall QoE Prediction}

\paragraph{Language Models}

After the dimension-level QoE prediction, we obtained the LLM-based datasets \texttt{TextQ-ATS-LLM} and \texttt{TextQ-MT-LLM} which contain an overall score as specified in Section~\ref{sec3.2:llm-corpora}, enabling the prediction of overall QoE. 

\begin{table}[htb]
\caption{Performance of the fine-tuned language models for overall QoE prediction on \texttt{TextQ-ATS-LLM}}
\centering
\begin{tabular}{lcccc}
\hline
    Model & RMSE & MAE & $R^2$ \\
\hline
\texttt{bert-base-german-uncased} & 1.1406 & 0.8721 & -0.0099 \\
\texttt{bert-base-german-cased} & 1.1558 & 0.8913 & -0.0502 \\
\texttt{gbert-base} & 1.1483 & 0.8638 & 0.0073 \\
\texttt{gbert-large} & \textbf{0.9502} & \textbf{0.7304} & \textbf{0.2859} \\
\texttt{gelectra-large} & 1.0045 & 0.7368 & 0.1920 \\
\hline
\end{tabular}
\label{tab:model_selection_ATS-LLM}
\end{table}


\begin{table}[htb]
\caption{Performance of the fine-tuned language models for overall QoE prediction on \texttt{TextQ-MT-LLM}}
\centering
\begin{tabular}{lcccc}
\hline
    Model & RMSE & MAE & $R^2$ \\
\hline
\texttt{bert-base-german-uncased} & 1.6404 & 1.3925 & -0.4543 \\
\texttt{bert-base-german-cased} & 1.3572 & 1.0988 & -0.0299 \\
\texttt{gbert-base} & 1.4740 & 1.2506 & -0.1061 \\
\texttt{gbert-large} & 1.1997 & 1.0241 & 0.1598 \\
\texttt{gelectra-large} & \textbf{1.1450} & \textbf{0.9929} & \textbf{0.3246} \\
\hline
\end{tabular}
\label{tab:model_selection_MT-LLM}
\end{table}

Tables~\ref{tab:model_selection_ATS-LLM} and~\ref{tab:model_selection_MT-LLM} report the results of fine-tuning language models for overall QoE prediction. Although these datasets were collected independently from the dimension-level QoE corpora, the same models performed best, namely \texttt{gbert-large} for ATS and \texttt{gelectra-large} for MT. We therefore used these again as the backbone models.

\paragraph{Joint Training}

\begin{table*}[h]
\caption{Performance of the fine-tuned language models for overall QoE prediction on \texttt{TextQ-ATS-LLM} and \texttt{TextQ-MT-LLM}}
\footnotesize
\centering
\begin{tabular}{lcccccc}
\hline
Model 
& \multicolumn{3}{c}{\texttt{TextQ-ATS-LLM}} 
& \multicolumn{3}{c}{\texttt{TextQ-MT-LLM}} \\
& RMSE & MAE & $R^2$ 
& RMSE & MAE & $R^2$ \\
\hline
\texttt{gbert-large} (ATS) 
& \textbf{0.9502} & \textbf{0.7304} & \textbf{0.2859} 
& -- & -- & -- \\

\texttt{gelectra-large} (MT) 
& -- & -- & --
& 1.1450 & 0.9929 & 0.3246   \\
\hline
\texttt{gbert-large} (ATS $\cup$ MT) 
& 1.2303 & 1.0197 & -2.4031
& 1.3214 & 1.1107 & -0.0089 \\

\texttt{gelectra-large} (ATS $\cup$ MT) 
& 0.9520 & 0.7358 & -0.6210
& \textbf{1.0066} & \textbf{0.8202} & \textbf{0.4362} \\
\hline
\texttt{gbert-large} (multi-task learning) 
& 1.0232 & 0.7567 & 0.1254 
& 1.2615 & 1.0348 & 0.1091 \\

\texttt{gelectra-large} (multi-task learning) 
& 0.9956 & 0.7584 & 0.1650
& 1.1613 & 0.9805 & 0.2962 \\

\hline
\end{tabular}
\label{tab:joint-ats-mt}
\end{table*}

Next, we explored whether jointly training on both ATS and MT datasets for overall QoE could improve performance. For each fold, we merged and shuffled the training, validation, and test splits from both datasets together. We evaluated two approaches: (1) standard fine-tuning on the merged dataset, and (2) multi-task learning with hard parameter sharing, where each dataset was treated as a separate task. Table~\ref{tab:joint-ats-mt} presents the performance of both approaches alongside the single-dataset baselines.

Joint training produced asymmetric results. For MT, \texttt{gelectra-large} fine-tuned on the merged dataset improved performance across all metrics. For ATS, however, no improvement was achieved. The best joint model (\texttt{gelectra-large} on merged data) closely matched the single-dataset baseline in RMSE (0.9520 vs. 0.9502), although its\(R^2\) was lower. Multi-task learning underperformed for both datasets.

We interpret these results as follows. The MT baseline had more room for improvement, with an RMSE of 1.1450 compared to 0.9502 for ATS. Joint training appears to have regularized the MT model effectively, leveraging additional data from the ATS corpus to improve generalization. Conversely, the ATS model could have already been near-optimal given its data. Adding MT data introduced noise or task interference, leading to marginal or negative effects. This pattern is consistent with observations in multi-task and transfer learning, where benefits occur primarily to tasks with higher baseline error or smaller datasets. The best-performing ATS model remained (\texttt{gbert-large} trained on ATS alone), while the best MT model became \texttt{gelectra-large} trained on the merged corpus. We therefore adopted these configurations as the backbones for the hybrid models in the next experiments.

\paragraph{Hybrid Models}

\begin{table}[h]
\caption{Performance of the hybrid models for overall QoE prediction on \texttt{TextQ-ATS-LLM}}
\centering
\begin{tabular}{lcccc}
\hline
    Model & RMSE & MAE & $R^2$ \\
\hline
\texttt{gbert-large} & 0.9502 & 0.7304 & 0.2859 \\
Hybrid language model & 0.9714 & 0.7514 & 0.2533 \\
Hybrid SVM & \textbf{0.8906} & \textbf{0.7106} & \textbf{0.3712} \\
\hline
\end{tabular}
\label{tab:hybrid-models-ats-overall}
\end{table}

\begin{table}[h]
\caption{Performance of the hybrid models for overall QoE prediction on \texttt{TextQ-MT-LLM}}
\centering
\begin{tabular}{lcccc}
\hline
    Model & RMSE & MAE & $R^2$ \\
\hline
\texttt{gelectra-large} (ATS $\cup$ MT) & \textbf{1.0066} & \textbf{0.8202} & \textbf{0.4362} \\
Hybrid language model (ATS $\cup$ MT) & 1.1726 & 0.9609 & 0.3206 \\
Hybrid SVM (ATS $\cup$ MT) & 1.0870 & 0.9262 & 0.3381 \\

\hline
\end{tabular}
\label{tab:hybrid-models-mt-overall}
\end{table}

Tables~\ref{tab:hybrid-models-ats-overall} and~\ref{tab:hybrid-models-mt-overall} compare the performance of the hybrid models against the previous baselines. For ATS, as was the case in the dimension-level experiments, the hybrid SVM outperformed the baseline, improving all three reported metrics. For MT, neither hybrid model surpassed the baseline.

These results demonstrate that we have successfully developed predictive models for QoE assessment of machine translation and text summarization, covering both fine-grained dimension-level scores and overall quality, while investigating a range of approaches to improve regression performance.

\subsection{LLM-Based Extension}
The \texttt{TextQ-ATS-LLM} and \texttt{TextQ-MT-LLM} datasets also include ratings for the identified QoE dimensions. We therefore extended the dimension-level prediction data by merging these with the original corpora and reran the language model and hybrid models on $\texttt{TextQ-ATS} \cup \texttt{TextQ-ATS-LLM}$ and $\texttt{TextQ-MT} \cup \texttt{TextQ-MT-LLM}$.

\begin{table}[h]
\caption{Performance of the models on $\texttt{TextQ-ATS} \cup \texttt{TextQ-ATS-LLM}$}
\centering
\begin{tabular}{lcccc}
\hline
    Model & RMSE & MAE & $R^2$ \\
\hline
Multi-Target (\texttt{gbert-large}) & 0.8988 & 0.7232 & 0.3636 \\
Hybrid language model & 0.9066 & 0.7220 & 0.3510 \\
Hybrid SVM & \textbf{0.8495} & \textbf{0.6844} & \textbf{0.4350} \\
\hline
\end{tabular}
\label{tab:ats-with-llm}
\end{table}

\begin{table}[h]
\caption{Performance of the models on $\texttt{TextQ-MT} \cup \texttt{TextQ-MT-LLM}$}
\centering
\begin{tabular}{lcccc}
\hline
    Model & RMSE & MAE & $R^2$ \\
\hline
Single-Target (\texttt{gelectra-large}) & 1.4836 & 1.2178 & -0.1254 \\
Hybrid language model & \textbf{1.4487} & 1.2066 & \textbf{-0.0744} \\
Hybrid SVM  & 1.4647 & \textbf{1.1964} & -0.0964 \\

\hline
\end{tabular}
\label{tab:mt-with-llm}
\end{table}

For ATS, prediction performance on the combined original and LLM-generated corpus was slightly lower across all models than on \texttt{TextQ-ATS} alone. One possible explanation is that the LLM-generated data exhibit distinct characteristics, making the combined dataset more heterogeneous and challenging. The hybrid SVM achieved the lowest error for ATS again. With an RMSE of 0.8495, the performance remains stable despite this marginal performance degradation. For MT, however, the extended dataset proved substantially more challenging. All models exhibited substantially worse performance compared to the non-LLM-based \texttt{TextQ-MT}. Despite this, these models will be evaluated on the held-out validation datasets, consisting of LLM-generated and non-LLM-based text samples.

\subsection{Final Evaluation}
Finally, we evaluated the best-performing models for dimension-level QoE and overall QoE. Table \ref{tab:final-val} presents the performance of the models on the respective validation sets.

\begin{table}[h]
\caption{Final validation performances on the QoE datasets}
\centering
\begin{tabular}{lcccc}
\hline
    NLG Task & QoE Label & RMSE & MAE & $R^2$ \\
\hline
ATS & Dimensions & 0.9197 & 0.7319 & 0.3968\\
ATS & Overall Score & 0.9572 & 0.7457 & 0.4353 \\
MT & Dimensions & 1.5833 & 1.2989 & -0.3177 \\
MT& Overall Score & 1.1824 & 0.9753 & 0.3358 \\
\hline
\end{tabular}
\label{tab:final-val}
\end{table}

Performance was generally lower on the held-out validation sets than in the prior development experiments. Yet, all metrics are relatively close to the performance the models demonstrated previously, speaking for the robustness of the dataset collection process and validating our QoE modeling. As was already indicated in the LLM-based extension experiments, dimension-level MT prediction remained the most challenging setting in the final evaluation. This may be caused by the greater heterogeneity of the MT data, which combines LLM- and non-LLM-based translations and therefore contains different surface-level error profiles than the original \texttt{TextQ-MT} corpus. Unlike ATS, where dimensions such as clarity, predictability, and linguistic logic are reflected in longer-range properties such as coherence, structure, and readability, MT outputs are mostly shorter and sentence-level, making the corresponding dimensions harder to distinguish from the text output cues alone. In particular, LLM translations may reduce obvious grammatical or readability errors, causing precision, transparency, complexity, and grammaticality to become less clearly separable. The stronger performance for the overall score suggests that MT QoE can be predicted more effectively at a coarser level, while fine-grained MT dimension prediction may require larger, more balanced data or alternative text features in future work.

\begin{figure}[htb]
    \centering
    \begin{minipage}{0.48\textwidth}
        \centering
        \includegraphics[width=\textwidth]{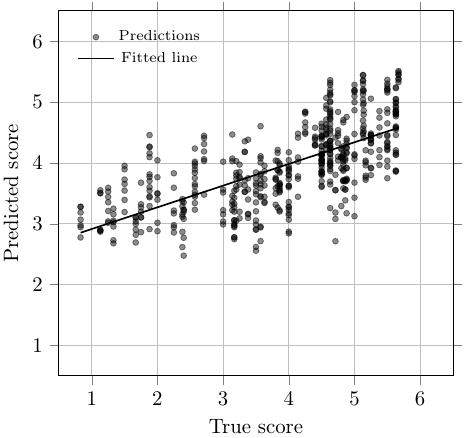}
        \caption{Scatter plot of the predictions on \texttt{TextQ-ATS-Val}}
        \label{fig:scatter-ats}
    \end{minipage}
    \hfill
    \begin{minipage}{0.48\textwidth}
        \centering
        \includegraphics[width=\textwidth]{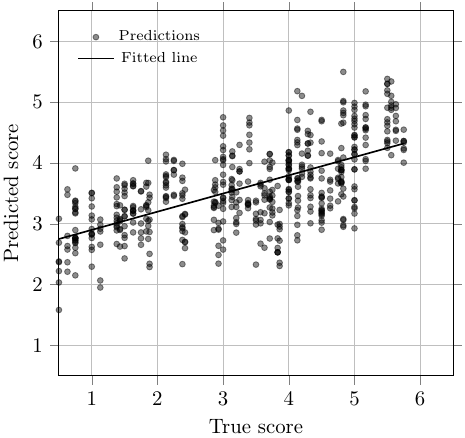}
        \caption{Scatter plot of the predictions on \texttt{TextQ-MT-Val}}
        \label{fig:scatter-mt}
    \end{minipage}
\end{figure}

The other validation sets yielded similar results to previous experiments. To interpret these results in absolute terms, recall that all QoE ratings lie on a 0 to 6 scale, where 0 represents the most negative and 6 the most positive perception. For ATS, an RMSE of approximately 0.92 to 0.96 is therefore below one point on this seven-point scale. Given the inherent subjectivity of human ratings and the complexity of the underlying quality dimensions, this level of accuracy is compelling. A MAE of around 0.73 to 0.75 further confirms that typical predictions deviate from human judgments by less than one scale point in absolute terms, which is also the case for overall MT QoE (0.98). The \(R^2\) values, ranging from 0.34 to 0.44, indicate that the models capture a substantial portion of the variance in human ratings. These values may be interpreted as meaningful and acceptable. This is plausible because QoE prediction is a human-centered perceptual task, where ratings depend on subjective judgments and annotator variability rather than deterministic text properties. In comparable human-centered domains, \(R^2\) values of 0.10–0.30 are often considered acceptable in social sciences and psychology \parencite{Ozili2022-ul}, while values above 0.15 have been described as meaningful in clinical research \parencite{Gupta2024-mh}. Considering this, our positive validation \(R^2\) values of 0.34–0.44 indicate reasonable explanatory capability for subjective QoE modeling. The scatter plots in Figures \ref{fig:scatter-ats} and \ref{fig:scatter-mt} further support this interpretation, as the fitted lines show a clear positive relationship between true and predicted overall QoE scores.

Overall, these results demonstrate the feasibility of automatic QoE prediction for German NLG and reveal differences across tasks and prediction targets. On the source-disjoint held-out sets, the models retain predictive capability for ATS dimension-level and overall QoE as well as overall MT QoE, whereas fine-grained MT dimension prediction remains challenging. These findings provide initial baselines for automatic QoE assessment using \texttt{TextQ-German} and identify directions for further modeling and data collection.

\section{Conclusion}

In this paper, we introduced \texttt{TextQ-German}, a novel dataset suite for Quality of Experience assessment of German Natural Language Generation, covering automatic text summarization and machine translation. Through crowdsourcing studies, we identified and validated four perceptual quality dimensions for each task: Precision, Complexity, Grammaticality, and Transparency for MT, and Linguistic Logic, Complexity, Clarity, and Predictability for ATS.

We developed and evaluated a range of automatic QoE prediction models, including transformer-based models, linguistic feature-based models, and hybrid approaches. Our experiments show that hybrid models, which combine neural embeddings with carefully selected interpretable linguistic features, improve over pure transformer-based baselines in most experiments. Notably, linguistic features alone, when selected through Sequential Feature Selection, achieve performance approaching that of fine-tuned language models, while offering the advantage of transparency, efficiency and interpretability.

The final validation on held-out sets comprising both LLM-generated and non-LLM-based texts provides evidence of out-of-sample generalization for ATS and overall MT QoE, while fine-grained MT dimension prediction remains more challenging.

As NLG systems become increasingly integrated into real-world applications, we believe that user-centered evaluation frameworks such as QoE will play an essential role in ensuring that generated text meets the expectations and needs of its human audience.

A key limitation of our study is that the QoE assessment is operationalized through quality attributes that are directly observable in the text (complexity, grammaticality, etc.). While these are central to the user's perception, a QoE evaluation may also involve contextual factors such as task utility, prior expectations, and emotional response, which we did not model. We therefore view our work as a novel resource and important step towards QoE assessment for NLG rather than the complete, finished realization of this. Furthermore, as obtaining human ratings is costly, the relatively small dataset size restricts the training of larger models and limits the generalizability.  Despite these limitations, the strong performance of hybrid models and the positive validation results demonstrate the utility of the resource.

For future work, several promising directions emerge. Extending \texttt{TextQ-German} to additional NLG tasks such as question-answering, dialogue generation, or image captioning would broaden the applicability of our QoE framework. Incorporating multilingual data would allow cross-lingual comparisons and help determine whether perceptual quality dimensions are language-universal or culturally dependent. Future work could explore more sophisticated multi-task and meta-learning architectures to better exploit the complementarity between ATS and MT datasets, potentially improving performance for tasks with limited data. While our linguistic features are hand-crafted and interpretable, automatically discovering novel features or leveraging large language models as feature extractors could further improve model performance. An additional methodological consideration for future work concerns the evaluation metrics themselves. Since human ratings exhibit natural variability, a prediction model should not be expected to outperform the agreement between human annotators. Metrics that account for this inherent uncertainty, such as a version of RMSE that considers predictions within the range of human standard deviation as effectively correct, would provide a more realistic assessment of model performance. Ultimately, advancing evaluation methods will be essential for developing NLG systems whose measured performance translates into meaningful improvements in the quality experienced by their users.

\FloatBarrier

\begin{appendix}
    
\section*{Appendix}

\section{Adjective Pairs}

\begin{table*}[!h]
\caption{Complete list of polar adjective pairs used in the experiments for the text type MT in the German original and translated into English for better understanding. Adapted from \textcite{Manakhimova2025_1255}}
\centering
\footnotesize
\begin{tabular}{p{0.28\textwidth}p{0.32\textwidth}p{0.32\textwidth}}
\toprule
& \textbf{German original} & \textbf{English translation} \\
\midrule

\textbf{Group 1:} final list of adjective pairs that are loading on the underlying factors, with the pairs used for the quality dimension quantification highlighted in boldface
&
\begin{tabular}[t]{@{}l@{}}
direkt -- umständlich \\
eindeutig -- mehrdeutig \\
\textbf{einfach -- kompliziert} \\
\textbf{grammatisch -- ungrammatisch} \\
klar -- wirr \\
\textbf{präzise -- ungenau} \\
\textbf{übersichtlich -- verwirrend} \\
vollständig -- lückenhaft
\end{tabular}
&
\begin{tabular}[t]{@{}l@{}}
direct -- ponderous \\
unambiguous -- ambiguous \\
\textbf{simple -- complicated} \\
\textbf{grammatical -- ungrammatical} \\
clear -- chaotic \\
\textbf{precise -- vague} \\
\textbf{neat -- confusing} \\
complete -- incomplete
\end{tabular}
\\
\midrule

\textbf{Group 2:} list of adjective pairs that were removed during the factor analysis for the sake of interpretability
&
\begin{tabular}[t]{@{}l@{}}
flüssig -- holprig \\
formell -- informell \\
geordnet -- durcheinander \\
geschrieben -- gesprochen \\
höflich -- unhöflich \\
kongruent -- inkongruent \\
konsistent -- inkonsistent \\
logisch -- unlogisch \\
menschlich -- technisch \\
muttersprachlich -- fremdsprachlich \\
persönlich -- unpersönlich \\
professionell -- laienhaft
\end{tabular}
&
\begin{tabular}[t]{@{}l@{}}
fluent -- non-fluent \\
formal -- informal \\
orderly -- messy \\
written -- spoken \\
polite -- impolite \\
congruent -- incongruent \\
consistent -- inconsistent \\
logical -- illogical \\
human -- technical \\
native -- foreign-language \\
personal -- impersonal \\
professional -- unprofessional
\end{tabular}
\\
\midrule

\textbf{Group 3:} list of adjective pairs that were removed after the preliminary study
&
\begin{tabular}[t]{@{}l@{}}
aktiv -- passiv \\
angemessen -- unangemessen \\
angenehm -- unangenehm \\
bedeutungsvoll -- bedeutungslos \\
bekannt -- unbekannt \\
förmlich -- lässig \\
gebildet -- ungebildet \\
gut -- schlecht \\
hochwertig -- minderwertig \\
informativ -- nichtssagend \\
kreativ -- simpel \\
lustig -- ernst \\
optimal -- suboptimal \\
praktisch -- unpraktisch \\
stilvoll -- stillos \\
vertraut -- fremd \\
vorhersehbar -- unberechenbar \\
warm -- kalt \\
weich -- hart \\
zweckorientiert -- zweckfrei
\end{tabular}
&
\begin{tabular}[t]{@{}l@{}}
active -- passive \\
appropriate -- inappropriate \\
pleasant -- unpleasant \\
meaningful -- meaningless \\
known -- unknown \\
formal -- casual \\
educated -- uneducated \\
good -- bad \\
valuable -- poor \\
informative -- bland \\
creative -- simple \\
funny -- serious \\
optimal -- suboptimal \\
practical -- impractical \\
classy -- unclassy \\
familiar -- foreign \\
predictable -- unpredictable \\
warm -- cold \\
soft -- hard \\
purposeful -- purposeless
\end{tabular}
\\
\bottomrule
\end{tabular}
\label{tab:mt_all_adjective_pairs}
\end{table*}

\begin{table*}[t]
\caption{Complete list of polar adjective pairs used in the experiments for the text type ATS in the German original and translated into English for better understanding. Adapted from \textcite{Manakhimova2025_1255}}
\centering
\footnotesize
\begin{tabular}{p{0.28\textwidth}p{0.32\textwidth}p{0.32\textwidth}}
\toprule
& \textbf{German original} & \textbf{English translation} \\
\midrule

\textbf{Group 1:} final list of adjective pairs that are loading on the underlying factors, with the pairs used for the quality dimension quantification highlighted in boldface
&
\begin{tabular}[t]{@{}l@{}}
\textbf{eindeutig -- mehrdeutig} \\
\\[-0.5em]
\textbf{einfach -- kompliziert} \\
\textbf{logisch -- unlogisch} \\
präzise -- ungenau \\
simpel -- komplex \\
vollständig -- lückenhaft \\
\textbf{vorhersehbar -- unberechenbar} \\
zusammenhängend -- unzusammen-\\
hängend
\end{tabular}
&
\begin{tabular}[t]{@{}l@{}}
\textbf{unambiguous -- ambiguous} \\
\\[-0.5em]
\textbf{simple -- complicated} \\
\textbf{logical -- illogical} \\
precise -- vague \\
straightforward -- complex \\
complete -- incomplete \\
\textbf{predictable -- unpredictable} \\
coherent -- incoherent
\end{tabular}
\\
\midrule

\textbf{Group 2:} list of adjective pairs that were removed during the factor analysis for the sake of interpretability
&
\begin{tabular}[t]{@{}l@{}}
anspruchsvoll -- anspruchslos \\
ausführlich -- knapp \\
direkt -- umständlich \\
fehlerfrei -- fehlerhaft \\
fließend -- holprig \\
geordnet -- ungeordnet \\
grammatisch -- ungrammatisch \\
klar -- wirr \\
sinnvoll -- sinnlos \\
übersichtlich -- verwirrend \\
verständlich -- unverständlich \\
widerspruchsfrei -- widersprüchlich \\
wortreich -- wortarm
\end{tabular}
&
\begin{tabular}[t]{@{}l@{}}
sophisticated -- unsophisticated \\
elaborate -- terse \\
direct -- ponderous \\
error-free -- faulty \\
fluent -- non-fluent \\
ordered -- disordered \\
grammatical -- ungrammatical \\
clear -- chaotic \\
meaningful -- meaningless \\
neat -- confusing \\
comprehensible -- incomprehensible \\
consistent -- contradictory \\
verbose -- concise
\end{tabular}
\\
\midrule

\textbf{Group 3:} list of adjective pairs that were removed after the preliminary study
&
\begin{tabular}[t]{@{}l@{}}
konsistent -- inkonsistent \\
tiefgehend -- oberflächlich \\
informativ -- uninformativ \\
nicht wiederholend -- wiederholend \\
geeignet -- unpassend \\
kongruent -- inkongruent \\
pragmatisch -- unpragmatisch \\
eigenständig -- uneigenständig \\
konkret -- abstrakt \\
kompatibel -- inkompatibel \\
normal -- skurril \\
beständig -- wechselhaft \\
prägnant -- ungenau \\
sinnig -- unsinnig \\
kohärent -- inkohärent \\
bekannt -- fremd \\
einheitlich -- uneinheitlich
\end{tabular}
&
\begin{tabular}[t]{@{}l@{}}
consistent -- inconsistent \\
thorough -- cursory \\
informative -- uninformative \\
non-repetitive -- repetitive \\
suitable -- unsuitable \\
congruent -- incongruent \\
pragmatic -- impragmatic \\
independent -- dependent \\
concrete -- abstract \\
compatible -- incompatible \\
normal -- eccentric \\
stable -- volatile \\
concise -- imprecise \\
meaningful -- nonsensical \\
coherent -- incoherent \\
familiar -- unfamiliar \\
uniform -- inconsistent
\end{tabular}
\\
\bottomrule
\end{tabular}
\label{tab:ats_all_adjective_pairs}
\end{table*}

\clearpage

\section{Generation Prompts}\label{app:prompts}
The following prompts were used to generate the LLM-based corpus extensions
described in Section~\ref{sec3.2:llm-corpora}.  Each source text was read from the
input data, inserted into the prompt at the position marked \texttt{<source text>},
and processed individually: the OpenAI models were called through the Chat Completions API, and the open-weight models through Ollama.

\paragraph{Machine translation (all models).}
\begin{lstlisting}[basicstyle=\ttfamily\small, breaklines=true, columns=fullflexible]
Translate the following English text to German:
<source text>
\end{lstlisting}

\paragraph{Summarization, locally hosted open-weight models (except SauerkrautLM).}
\begin{lstlisting}[basicstyle=\ttfamily\small, breaklines=true, columns=fullflexible]
Fasse den folgenden Text auf Deutsch so kurz wie möglich zusammen.
<source text>
Kurze Zusammenfassung (Deutsch):
\end{lstlisting}

\paragraph{Summarization, SauerkrautLM-7B-v1 (Alpaca format).}
\begin{lstlisting}[basicstyle=\ttfamily\small, breaklines=true, columns=fullflexible]
### Anweisung:
FASS DEN FOLGENDEN TEXT PRÄGNANT IN 3-5 SÄTZEN ZUSAMMEN. KONZENTRIERE DICH AUF
DIE HAUPTAUSSAGEN UND WICHTIGSTEN PUNKTE. VERMEIDE WIEDERHOLUNGEN UND
NEBENSÄCHLICHKEITEN.
### Eingabe:
<source text>
### Antwort:
\end{lstlisting}

\paragraph{Summarization, GPT-4o length-constrained variant (gpt-4o\_short).}
\begin{lstlisting}[basicstyle=\ttfamily\small, columns=fullflexible]
Fasse den folgenden Text sehr prägnant in maximal 3 sehr kurzen
Sätzen zusammen. Verwende einfache und kurze Formulierungen.
Halte jeden Satz möglichst unter 15 Wörtern.
Die Zusammenfassung soll auf Deutsch sein:
<source text>
\end{lstlisting}

\section{Linguistic Features}
\begin{table*}[h!]
\centering
\footnotesize
\caption{All features used in the feature selection process, categorized by feature type}
\begin{tabularx}{\textwidth}{>{\raggedright\arraybackslash}p{0.25\textwidth} >{\raggedright\arraybackslash}X}\hline
Feature type & Features \\ \hline

Readability features & Gunning Fog Index, Flesch reading ease, SMOG index, Coleman-Liau index, adjective ttr, adposition ttr, adverb ttr, aux ttr, Wiener Sachtextformel 1-4\\ \hline

Lexical richness features & Dugast's Uber Index, Yule's K, Simpson's D, Herdan's lexical diversity measure, Summer's lexical diversity measure, Maas's lexical diversity measure, lexical density, noun variation, content word distribution, adjective variation, adverb variation, modifier variation, verb variation, type token ratio, root type token ratio, average sentence length in words, average word length in characters, average word length in syllables, unique tokens count, bilog ttr, conjunction ttr, cttr,  determiner ttr, grammatical item ttr, lexical item ttr, lexical verb ttr, noun ttr, particle ttr, pronoun ttr, proper noun ttr, verb ttr, zipf goodness of fit, zipf steepness of curve\\ \hline

Syntactic features & Noun phrases per sentence count, verb phrases per sentence count, average sentence length in characters, characters per sentence count, finite verbs count, foreign words count, infinitive verb count, noun to pronoun ratio, passive voice count, past tense count, present tense count, pronouns per sentence count, punctuation count,  standard deviation tokens per sentence, verb to adverb ratio, verb to noun ratio, verbs per sentence count\\\hline

Morphological features & Infinitive verbs to verbs ratio, participle verbs to verbs ratio, nominative nouns to nouns ratio, genitive nouns to nouns ratio, dative nouns to nouns ratio, accusative nouns to nouns ratio, haben to verb ratio, sein to verb ratio, noun token ratio, verb token ratio, finite verbs to verbs ratio, first person verbs to finite verbs ratio, second person verbs to finite verbs ratio, third person verbs to finite verbs ratio, imperative verbs to verbs ratio, particle verbs to verbs ratio, past tense verbs to finite verbs ratio, present tense verbs to finite verbs ratio,  subjunctive verbs to finite verbs ratio, percentage of words with 6 or more letters, adjective count, adposition count, adverb count, aux to verb ratio, aux verb count, conjunction count, definite word count, demonstrative count, determiners count, first person pronoun count, grammatical item count, indefinite word count, interrogative count, lexical item count, noun count, numeral count, particle count,  personal pronoun count, plural word count, proper noun count, second person pronoun count, singular word count, verb count, 
average adjective length, average adposition length, average adverb length, average aux length, average conjunction length, average determiner length, average lexical verb length, average noun length, average numeral length, average particle length, average pronoun length, average proper noun length, average verb length,  percentage of monosyllabic words, percentage of words with 3 or more syllables \\\hline
\label{tab:all-features}
\end{tabularx}
\end{table*}

\begin{table*}[!t]
\centering
\caption{Performance of feature selection methods on \texttt{TextQ-ATS} by quality dimension}
\label{tab:features_ats_detailed}

\footnotesize
\setlength{\tabcolsep}{2pt}
\renewcommand{\arraystretch}{0.78}

\begin{tabular}{@{}p{1.7cm}p{1cm}p{0.9cm}p{1cm}p{7.5cm}@{}}
\toprule
Dimension & Method & RMSE & \# Feat. & Selected features \\
\midrule
Logic 
& RFE & 1.2279 & 1
& yules k \\

& SFS & \textbf{1.1316} & 20
& average number syllables, smog index, adposition count, particle count, first person pronoun count, second person pronoun count, interrogative pronoun count, indefinite word count, avg aux length, noun variation, first person verbs to finite verbs ratio, imperative verbs to verbs ratio, second person verbs to finite verbs ratio, third person verbs to finite verbs ratio, subjunctive verbs to finite verbs ratio, std dev tokens per sentence, avg adposition length, avg determiner length, avg numeral length, adposition ttr \\

& Lasso & 1.1997 & 1
& indefinite word count \\

& Elastic & 1.1997 & 1
& indefinite word count \\

\midrule
Complexity
& RFE & 1.1347 & 1
& noun count \\

& SFS & \textbf{0.8336} & 20
& smog index, type token ratio, number unique tokens, content word distribution, numerical count, first person pronoun count, demonstrative count, definite word count, present tense count, avg aux length, first person verbs to finite verbs ratio, imperative verbs to verbs ratio, present tense verbs to finite verbs ratio, past tense verbs to finite verbs ratio, second person verbs to finite verbs ratio, third person verbs to finite verbs ratio, subjunctive verbs to finite verbs ratio, std dev tokens per sentence, noun to pronoun ratio, conjunction ttr \\

& Lasso & 0.9340 & 22
& wiener sachtextformel 1, root type token ratio, corrected type token ratio, lexical density, content word distribution, verb count, aux verb count, numerical count, demonstrative count, present tense count, noun token ratio, noun phrases per sentence count, finite verbs to verbs ratio, std dev tokens per sentence, pronouns per sentence count, avg numeral length, noun ttr, lexical items ttr, adposition ttr, aux ttr, conjunction ttr, grammatical ttr \\

& Elastic & 1.0582 & 28
& wiener sachtextformel 1, wiener sachtextformel 2, wiener sachtextformel 3, wiener sachtextformel 4, type token ratio, root type token ratio, corrected type token ratio, lexical density, number unique tokens, content word distribution, aux verb count, numerical count, demonstrative count, past tense count, present tense count, adjective variation, noun token ratio, noun phrases per sentence count, participle verbs to verbs ratio, finite verbs to verbs ratio, std dev tokens per sentence, pronouns per sentence count, avg numeral length, noun ttr, lexical items ttr, adposition ttr, aux ttr, grammatical ttr \\

\midrule
Clarity
& RFE & 0.9368 & 1
& yules k \\

& SFS & \textbf{0.8211} & 20
& adposition count, particle count, first person pronoun count, second person pronoun count, interrogative pronoun count, demonstrative count, singular word count, present tense count, first person verbs to finite verbs ratio, imperative verbs to verbs ratio, second person verbs to finite verbs ratio, std dev tokens per sentence, pronouns per sentence count, avg proper noun length, avg adposition length, avg determiner length, avg numeral length, lexical items ttr, adposition ttr, conjunction ttr \\

& Lasso & -- & 0
& -- \\

& Elastic & -- & 0
& -- \\

\midrule
Predictability
& RFE & 0.9098 & 1
& yules k \\

& SFS & \textbf{0.7828} & 20
& conjunction count, first person pronoun count, second person pronoun count, interrogative pronoun count, demonstrative count, singular word count, present tense count, adjective variation, first person verbs to finite verbs ratio, imperative verbs to verbs ratio, present tense verbs to finite verbs ratio, past tense verbs to finite verbs ratio, second person verbs to finite verbs ratio, third person verbs to finite verbs ratio, subjunctive verbs to finite verbs ratio, std dev tokens per sentence, pronouns per sentence count, avg proper noun length, lexical items ttr, conjunction ttr \\

& Lasso & 0.8866 & 1
& present tense count \\

& Elastic & 0.8866 & 1
& present tense count \\

\bottomrule
\end{tabular}

\vspace{0.3em}
\begin{minipage}{0.98\textwidth}
\tiny
The best RMSE within each quality dimension is highlighted in bold. A dash indicates that the method selected no features and therefore produced no valid RMSE.
\end{minipage}
\end{table*}

\begin{table*}[!t]
\centering
\caption{Performance of feature selection methods on \texttt{TextQ-MT} by quality dimension}
\label{tab:features_mt_detailed}

\footnotesize
\setlength{\tabcolsep}{2pt}
\renewcommand{\arraystretch}{0.78}

\begin{tabular}{@{}p{1.9cm}p{1cm}p{0.85cm}p{1.1cm}p{7.5cm}@{}}
\toprule
Dimension & Method & RMSE & \# Feat. & Selected features \\
\midrule
Complexity
& RFE & 1.1028 & 13
& coleman liau index, average chars per word, type token ratio, root type token ratio, corrected type token ratio, herdans measure, summers measure, maas measure, yules k, simpsons d, number unique tokens, verb count, aux verb count \\

& SFS & \textbf{0.8655} & 20
& average sentence length, number unique tokens, numerical count, interrogative pronoun count, past tense count, passive voice count, participle verbs to verbs ratio, imperative verbs to verbs ratio, accusative nouns to nouns ratio, dative nouns to nouns ratio, genitive nouns to nouns ratio, std dev tokens per sentence, verb to adverb ratio, characters per sentence count, avg sentence length in characters, avg conjunction length, lexical verb ttr, verb ttr, conjunction ttr, determiner ttr \\

& Lasso & 0.9291 & 13
& average sentence length, type token ratio, numerical count, interrogative pronoun count, past tense count, haben to verb ratio, participle verbs to verbs ratio, dative nouns to nouns ratio, genitive nouns to nouns ratio, characters per sentence count, avg sentence length in characters, avg conjunction length, verb ttr \\

& Elastic & 0.9630 & 15
& average sentence length, flesch reading ease, type token ratio, lexical density, numerical count, interrogative pronoun count, past tense count, haben to verb ratio, participle verbs to verbs ratio, dative nouns to nouns ratio, genitive nouns to nouns ratio, characters per sentence count, avg sentence length in characters, avg conjunction length, verb ttr \\

\midrule
Grammaticality
& RFE & 1.3810 & 1
& simpsons d \\

& SFS & \textbf{1.0656} & 20
& flesch reading ease, maas measure, number unique tokens, adposition count, numerical count, present tense count, bilogarithmic ttr, sein to verb ratio, haben to verb ratio, verb phrases per sentence count, first person verbs to finite verbs ratio, participle verbs to verbs ratio, past tense verbs to finite verbs ratio, third person verbs to finite verbs ratio, genitive nouns to nouns ratio, std dev tokens per sentence, avg adposition length, avg determiner length, adjective ttr, determiner ttr \\

& Lasso & 1.2994 & 2
& flesch reading ease, numerical count \\

& Elastic & 1.3105 & 3
& average sentence length, flesch reading ease, numerical count \\

\midrule
Transparency
& RFE & 1.2641 & 13
& average sentence length, flesch reading ease, type token ratio, herdans measure, summers measure, maas measure, yules k, simpsons d, number unique tokens, bilogarithmic ttr, zipf steepness curve, avg determiner length, determiner ttr \\

& SFS & \textbf{1.1611} & 20
& percentage words three syllables, flesch reading ease, root type token ratio, corrected type token ratio, grammatical item count, numerical count, plural word count, past tense count, passive voice count, participle verbs to verbs ratio, imperative verbs to verbs ratio, third person verbs to finite verbs ratio, genitive nouns to nouns ratio, verb to adverb ratio, characters per sentence count, avg sentence length in characters, avg noun length, avg particle length, noun ttr, determiner ttr \\

& Lasso & 1.2579 & 6
& flesch reading ease, lexical item count, numerical count, participle verbs to verbs ratio, dative nouns to nouns ratio, genitive nouns to nouns ratio \\

& Elastic & 1.2761 & 8
& average sentence length, flesch reading ease, lexical item count, numerical count, haben to verb ratio, participle verbs to verbs ratio, dative nouns to nouns ratio, genitive nouns to nouns ratio \\

\midrule
Precision
& RFE & 1.2975 & 1
& yules k \\

& SFS & \textbf{1.0106} & 20
& percentage words six letters, flesch reading ease, type token ratio, corrected type token ratio, herdans measure, maas measure, adposition count, numerical count, demonstrative count, definite word count, bilogarithmic ttr, verb phrases per sentence count, infinitive verbs to verbs ratio, participle verbs to verbs ratio, third person verbs to finite verbs ratio, nominative nouns to nouns ratio, std dev tokens per sentence, verb to adverb ratio, avg determiner length, determiner ttr \\

& Lasso & 1.2428 & 1
& numerical count \\

& Elastic & 1.2428 & 1
& numerical count \\

\bottomrule
\end{tabular}

\vspace{0.3em}
\begin{minipage}{0.98\textwidth}
\tiny
The best RMSE within each quality dimension is highlighted in bold.
\end{minipage}
\end{table*}

\end{appendix}

\FloatBarrier

\backmatter

\bmhead{Acknowledgements}
We thank Dominik Frefel, Manfred Vogel, and Fabian Märki for providing the GeWiki summary corpus they developed for the German Text Summarization Challenge of SwissText \& KONVENS 2020. We are also grateful to our colleagues for participating in our pre-study. We express our gratitude to Aleksandra Gabryszak for her insights into automatic text summarization and, together with Polina Danilovskaia, for their valuable support with dataset quality control and the  training of an exploratory, preliminary model.

The presented research was supported by the Deutsche Forschungsgemeinschaft (DFG) through the project “Analyse und automatische Abschätzung der Qualität maschinell generierter Texte”, project number 436813723.

\section*{Declarations}

\bmhead{Funding}
The presented research was funded by the Deutsche Forschungsgemeinschaft (DFG) through the project “Analyse und automatische Abschätzung der Qualität maschinell generierter Texte”, project number 436813723.

\bmhead{Conflict of interest}
The authors declare no Conflict of interest.

\bmhead{Ethics approval and consent to participate}
Not applicable.

\bmhead{Consent for publication}
Not applicable.

\bmhead{Data availability}
Our experiment code, adjective pairs, textual/linguistic features, and datasets can be accessed at \url{https://github.com/DFKI-NLP/TextQ/}. The dataset is additionally available at \url{https://huggingface.co/datasets/nphamdinh/textq-german} and \url{https://www.kaggle.com/datasets/namphamdinh/textq-german/}.

\bmhead{Materials availability}
Not applicable.

\bmhead{Code availability}
Available at \url{https://github.com/DFKI-NLP/TextQ/}.

\bmhead{Author contributions}
Drafting of the manuscript, conceptualization and implementation of the prediction experiments, analysis and discussion of the model results were led by Dinh Nam Pham.
Shushen Manakhimova and Vivien Macketanz jointly
conceptualized and compiled the datasets, including the design and execution of the crowdsourcing studies, quality dimension identification and statistical analysis, and the LLM-based corpus
extensions. Sebastian Möller provided supervision and acquired funding for the project.
All authors read and approved the final manuscript.

\printbibliography

\end{document}